\documentclass[10pt,twocolumn,letterpaper]{article}
\usepackage[pagenumbers]{cvpr}
\makeatletter
\@ifpackageloaded{lineno}{%
  \IfHookEmptyTF{build/column/before}{}{\RemoveFromHook{build/column/before}[lineno]}%
  \AddToHook{build/column/after}{%
    \setbox\@outputbox\vbox{%
      \boxmaxdepth\@maxdepth
      \protected@write\@auxout{}{%
        \string\def\string\@LN@column{\if@firstcolumn1\else2\fi}}%
      \box\@outputbox}}%
}{}
\makeatother
\usepackage{microtype}
\usepackage{graphicx}
\usepackage{amsmath}
\usepackage{amssymb}
\usepackage{booktabs}
\usepackage{caption}
\usepackage{placeins}
\usepackage[pagebackref]{hyperref}
\hypersetup{hidelinks,
  pdftitle={Beyond the Training Cameras: Coverage-Controlled Probes of Viewing Direction in Frozen Vision Models},
  pdfauthor={Peize Ding}}

\def\paperID{27673}
\def\confName{CVPR}
\def\confYear{2027}

\newcommand{\ua}{\mathbf{u}}
\newcommand{\R}{\mathbb{R}}
\newcommand{\supp}[1]{Sec.~S#1}

\title{Beyond the Training Cameras: Coverage-Controlled Probes\\of Viewing Direction in Frozen Vision Models}

\author{Peize Ding\\
Columbia University\\
{\tt\small pd2714@columbia.edu}
}

\begin{document}
\twocolumn[{%
\renewcommand\twocolumn[1][]{#1}%
\maketitle
\begin{center}
  \centering
  \includegraphics[width=\textwidth]{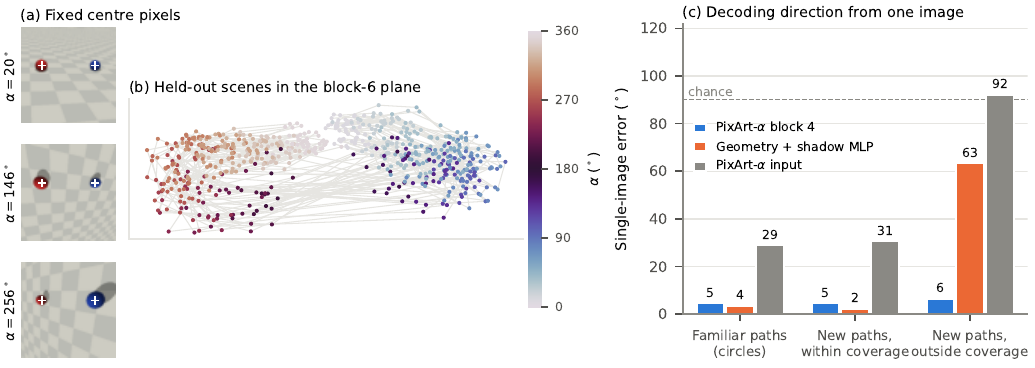}
  \captionof{figure}{\textbf{Viewing direction in a fixed-projection scene.}
  (a) Three views of one scene along a circular camera path, labelled by the viewing
  direction $\alpha$; white crosses mark the projected object centres, which are the
  same pixels in every view. (b) Trajectory-centred difference features of the test
  scenes at block 6 of PixArt-$\alpha$ ($t=200$), projected onto the plane fitted on
  training scenes (Eq.~\ref{eq:plane}); colour: $\alpha$; grey lines connect the views
  of one trajectory in order of $\alpha$. (c) Mean angular error of $\alpha$ decoded from
  one image of a test scene by a linear readout of PixArt-$\alpha$ block 4 (blue), a multilayer
  perceptron (MLP) on measured object and shadow geometry (orange) and a linear readout of the
  PixArt-$\alpha$ input (grey), on the training paths and on new paths within and outside
  the coverage of the training cameras (Sec.~\ref{sec:coverage}); dashed line: chance.}
  \label{fig:teaser}
\end{center}
\vspace{0.5em}
}]

\begin{abstract}
Vision models are trained on 2D images of a 3D world. Do their hidden features encode
the direction from which the camera views a scene, and if so, in what form? We study
the relative viewing direction $\alpha$ of two objects in rendered scenes where both
object centres stay at fixed pixels while the camera moves, so that perspective,
shading and shadows change but the projected layout does not. In eight pretrained
models, from diffusion models to self-supervised and image--text encoders, $\alpha$ is
encoded in a strikingly simple form: a single two-dimensional plane in the hidden
space, shared across scenes, in which the difference between the two objects'
features traces $(\cos\alpha,\sin\alpha)$. The plane keeps its orientation when
lighting, colours, floors and object shapes change. This encoding is more than a
summary of a few visible measurements: readouts fitted on one set of camera paths stay
accurate at camera positions outside their range in seven of the eight models,
whereas decoders of the measured object and shadow geometry, equally accurate within
that range, fail there. In the diffusion transformer PixArt-$\alpha$, the plane forms
abruptly at a single block and carries direction information to later blocks, yet
editing it changes the objects rather than the generated viewpoint. We also examine
how this encoding carries over to scanned real objects.
\end{abstract}

\section{Introduction}
\label{sec:intro}

Vision models are trained on two-dimensional images, yet the scenes in those images
are three-dimensional. Whether a model builds an internal description of the 3D scene,
such as where the camera stands relative to the objects, or only learns regularities of
the 2D image, is a basic question about what these models learn. It also matters in
practice: a generator that keeps track of the viewpoint can render objects, shadows and
perspective consistently and could be steered to a new camera position. The evidence
so far is mixed. Image generators produce objects, shadows and perspective that agree
with a single viewpoint~\cite{peebles2023dit,chen2024pixart}, and depth, surface
orientation and correspondence can be read from the internal features of many
models~\cite{chen2023beyond,zhan2024general,bhattad2023stylegan,tang2023dift,elbanani2024probing}.
Yet generators still make geometric mistakes~\cite{sarkar2024shadows}, and arranging
objects correctly in space remains a weakness of text-to-image
models~\cite{huang2023t2icompbench,ghosh2023geneval}.

We study one basic 3D variable, the direction from which the camera views a pair of
objects, and ask whether and how vision models encode it in their hidden features.
Viewing direction is a property of the camera and the scene as a whole, not of any
single object's appearance, so a model that represents it holds a piece of 3D scene
structure. Beyond its presence, we ask how it is encoded: at which layer it appears,
whether it forms a two-dimensional representation (a plane) in the hidden space shared
across scenes, whether later layers and
the generation process use it, and whether it extends to camera positions not seen
when the probe was fitted.

The standard tool for such questions is \emph{probing}: a simple predictor, usually a
linear map, is trained to read a property from the model's internal features, and its
success is taken as evidence that the model represents the property. For viewing
direction, probing alone has difficulty separating a 3D variable from 2D image cues.
The images used to train and test a probe are usually taken from the same range of
camera positions, and within that range the direction can be worked out from a few
quantities visible in the image, such as how large each object appears and where its
shadow falls. A probe that succeeds there may be reading these simple measurements
rather than a notion of direction held by the model, and a predictor that takes the
measurements themselves as input would do just as well. Moreover, a flexible enough
probe can extract information that the model never organises or
uses~\cite{hewitt2019control,belinkov2022probing}.

We design a method that answers these questions while avoiding these pitfalls. First,
a rendered dataset fixes the image positions of both object centres while the camera
moves, so the projected layout is constant and direction must be inferred from
perspective, shading, shadows and ground appearance (Fig.~\ref{fig:teaser}a). Second,
besides probes that decode $\alpha$, we test whether $\alpha$ forms a plane in the hidden
space: a single linear map, shared across scenes, must predict how held-out features
change with $(\cos\alpha,\sin\alpha)$ (Fig.~\ref{fig:teaser}b). Third, to
separate a 3D variable from image measurements, we evaluate probes on new camera paths
and split them by \emph{coverage}: whether the camera lies within the range of the
training cameras, where following a new path is interpolation, or outside it, where
the probe must extrapolate (Fig.~\ref{fig:teaser}c). Fourth, to ask whether the model
uses the encoding, we intervene on it in a diffusion transformer during the forward
pass and during generation. We apply this analysis to eight pretrained models, kept frozen (not fine-tuned), with
different training objectives and to datasets of increasing appearance variation, up
to scanned real objects.

Our contributions are:
\begin{itemize}
\item A fixed-projection benchmark for relative viewing direction, with shape,
appearance and photo-realistic variants of the same cameras, and a coverage-controlled
protocol that separates interpolation from extrapolation of camera position.
\item The finding that all eight models encode viewing direction as a single plane in
their hidden space, shared across scenes, in which the difference between the two
objects' features traces $(\cos\alpha,\sin\alpha)$. Decoding does not require this,
since each scene could use its own directions; the plane is strongest in the
diffusion transformers and DINOv2.
\item Evidence that linear readouts of seven of eight frozen models, generative or not,
extrapolate to camera positions outside the training range, where decoders of the
visible geometry fail.
\item A mechanistic analysis of PixArt-$\alpha$: the plane is written at one block and
carries direction information downstream, but it is read at the objects, so neither
clamping it nor gradient guidance rotates the generated viewpoint.
\item An analysis of shape and appearance: the plane keeps its orientation across
factors, extrapolation fails for shapes with elevation-dependent cues, and with real
objects each pair carries a plane of its own.
\end{itemize}

\section{Related Work}
\label{sec:related}

\paragraph{Scene properties in frozen models.}
Internal features of latent diffusion models~\cite{rombach2022ldm} linearly encode
depth and salient-object information~\cite{chen2023beyond}, support 3D scene
property probes~\cite{zhan2024general}, and yield dense correspondence without
supervision~\cite{tang2023dift}. Generative adversarial networks encode intrinsic
images~\cite{bhattad2023stylegan} and interpretable units~\cite{bau2019gandissection}.
Probe3D compares foundation models on depth, normals and multi-view
consistency~\cite{elbanani2024probing}. We study a scene-level variable, ask whether it
occupies a shared subspace, and test probes outside the range of training viewpoints.

\paragraph{Linear representations and probing methodology.}
Linear probes~\cite{alain2016probes} are informative only with appropriate controls,
since expressive probes can decode properties the model does not
use~\cite{hewitt2019control,belinkov2022probing}. In language models, spatial and
temporal coordinates are linearly represented~\cite{gurnee2024space}, and the linear
representation hypothesis has been formalised~\cite{park2024linear}. Our shared-plane
probe, which predicts features from the label with two directions common to all
scenes, constrains the representation more than a decoder, and the coverage split adds
a control that image measurements cannot pass.

\paragraph{Interventions and steering.}
Causal mediation~\cite{vig2020causal} and causal tracing~\cite{meng2022rome} test
whether representations are used; projecting out a probe subspace tests whether
behaviour depends on it~\cite{ravfogel2020inlp,elazar2021amnesic}, and activation
addition steers outputs~\cite{turner2023activation}. In image generators, latent
directions control pose and other
attributes~\cite{shen2020interfacegan,harkonen2020ganspace,kwon2023hspace}; diffusion
editing uses noising~\cite{meng2022sdedit}, deterministic
inversion~\cite{song2021ddim,mokady2023nulltext,hertz2023p2p} or classifier
guidance~\cite{dhariwal2021diffusion}. Unlike pose-conditioned novel-view
synthesis~\cite{liu2023zero123}, we ask whether an unconditioned model carries an
internal viewing-direction variable.

\section{Experimental Design}
\label{sec:method}

Our design has two aims: to make viewing direction the only thing that changes between
views of a scene, and to test whether a model's encoding of it reaches beyond the
camera positions seen during fitting.

\subsection{Fixed-projection dataset}

\begin{figure*}[!t]
  \centering
  \includegraphics[width=\textwidth]{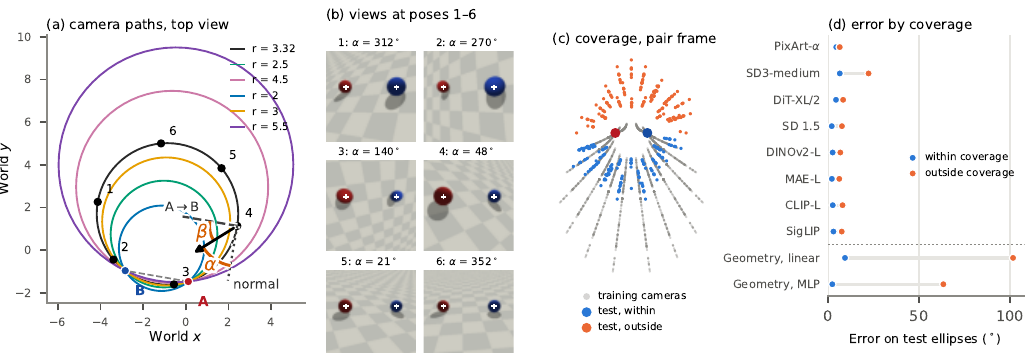}
  \caption{\textbf{Fixed-projection camera trajectories and coverage.} (a) Top view of
  the six circular camera paths around spheres A and B in one scene ($r$: radius). At
  pose 4, the arrow is the ground projection of the camera's optical axis, the dashed
  line the direction A$\rightarrow$B and the dotted line the normal of the pair;
  $\beta$ is the angle from A$\rightarrow$B to the axis and $\alpha=90^\circ-\beta$ the
  angle between the axis and the normal. (b) Rendered views at the numbered poses;
  white crosses mark the projected object centres, which are the same pixels in every
  image. (c) Camera positions of all scenes in the object-pair frame; red and blue
  discs: objects A and B; grey: training cameras; blue: test views within coverage
  (200 of 448); orange: test views outside coverage (203). (d) Mean single-image error
  on test ellipse views of spheres within (blue) and outside (orange) coverage, for the
  readout of each model and for the geometric decoders.}
  \label{fig:dataset}
\end{figure*}

If the objects moved in the image as the camera moved, direction could be read off
their positions. Each scene therefore contains a red sphere A and a blue sphere B of
equal diameter on a checkered ground plane with a fixed point light and cast shadows
(Fig.~\ref{fig:dataset}a,b), and for every camera pose the renderer solves for focal
length and camera orientation so that the centres of A and B project to the same two
pixels of a $256\times256$ image, to numerical precision (\supp{1}). Cameras lie on six
circles whose ground-plane projections pass through both object centres and on four
ellipses centred on the midpoint of the pair. The 48 scenes differ in object
separation and diameter, and each trajectory contributes 14 views. The label $\alpha$
is the horizontal viewing direction relative to the normal of the object pair:
$\alpha=(\pi/2-\beta)\bmod 2\pi$, where $\beta$ is the signed angle from the world
direction A$\rightarrow$B to the ground projection of the camera's optical axis, so
$\alpha=0$ when the camera looks along the normal of the pair (Fig.~\ref{fig:dataset}a,
pose 4). Because both centre
projections are fixed, $\alpha$ is not a function of the projected layout; it changes
apparent size, shading, shadows and ground appearance (Fig.~\ref{fig:teaser}a).

\subsection{Shape and appearance variants}

To find which image cues a readout relies on, we re-render the same views with one
factor changed at a time, and then with all factors varied and with real objects
(Fig.~\ref{fig:realism}). Cameras and object centres are unchanged, so labels, splits
and noise seeds are identical across variants. In \emph{Tier A}, every scene contains
a red and a blue object of one type and only the camera moves: spheres (the main
dataset), cylinders, or cubes whose faces are aligned with the A--B axis
(\emph{pair-aligned}) or with the world axes (\emph{world-aligned}; since the pair has
a random orientation in each scene, their appearance depends on $\alpha$ plus a
scene-specific offset). \emph{Tier B} keeps the spheres and varies the light position
(B1), the object colours (B2) or the floor (B3) across scenes (fixed within a scene); in B4, shapes, colours,
floor and light are all drawn at random (B5, with held-out shapes and colours, is in
\supp{9}). In \emph{Tier C}, each scene contains two scanned household objects from
Google Scanned Objects~\cite{downs2022gso}, each used once, randomly rotated and
centred on the original object centres, lit by a Poly Haven high-dynamic-range
environment map on a textured floor and path-traced with
Mitsuba~3~\cite{jakob2022mitsuba3}.

\subsection{Models and features}

To test whether the results depend on generative training, we probe eight frozen
models with three kinds of objective: the text-to-image diffusion transformers (DiTs)
PixArt-$\alpha$ XL/2~\cite{chen2024pixart} and Stable Diffusion 3
(SD3-medium)~\cite{esser2024sd3}, the class-conditional DiT-XL/2~\cite{peebles2023dit}
(null class) and the Stable Diffusion 1.5 (SD~1.5) UNet~\cite{rombach2022ldm}; the
self-supervised vision transformers (ViTs) DINOv2 ViT-L/14~\cite{oquab2024dinov2} and
masked autoencoder (MAE) ViT-L/16~\cite{he2022mae} (masking disabled); and the
image--text encoders CLIP ViT-L/14-336~\cite{radford2021clip} and SigLIP
So400m/14~\cite{zhai2023siglip}. Diffusion models receive the latent of the image under
their variational autoencoder, noised by the forward process to diffusion timestep
$t\in\{50,\dots,800\}$ of 1{,}000 (for SD3, the rectified-flow noise level with the
same signal-to-noise ratio), with the same noise sample for all views of a scene and a
fixed prompt (\supp{3}). Encoders receive clean images at their native resolution
(\supp{2}).

We record the input embedding (the tokens before the first block) and every block
output. Each object's feature,
$\mathbf a$ for A and $\mathbf b$ for B, averages the tokens of a fixed
$18\times18$-pixel square at its projected centre; the square lies inside the
silhouette in every view, so the feature describes the object alone. For Tier C, where
thin real objects leave no such square, we pool fixed windows of 8, 16 or 24 pixels
that may include background. We analyse the concatenation $\mathbf h=[\mathbf a;\mathbf
b]$, which keeps each object's information, and the difference $\mathbf h=\mathbf a-\mathbf b$,
which keeps only what differs between the objects; $d$ denotes the dimension of
$\mathbf h$.

\subsection{Shared-plane probe}
\label{sec:plane}

A decoder can recover $\alpha$ however it is stored. We test something more specific:
whether $\alpha$ forms a two-dimensional representation, a plane, in the hidden space,
the same plane for every scene. For each scene, trajectory and noise seed,
we subtract the mean feature over the trajectory's views, giving $\tilde{\mathbf h}$;
this removes what is constant along a trajectory, such as object size and appearance,
and leaves the part that changes with the view. We then fit
\begin{equation}
  \tilde{\mathbf h}(\alpha)\approx W\,\ua(\alpha),\qquad
  \ua(\alpha)=\begin{bmatrix}\cos\alpha\\ \sin\alpha\end{bmatrix},\qquad
  W\in\R^{d\times2},
  \label{eq:plane}
\end{equation}
with one $W$ shared by all scenes, by ridge regression (penalty $10^{-3}$). The column
span of $W$ is the \emph{plane}. We report held-out $R^2$: the fraction of the variance
of the centred features of test scenes, summed over all $d$ dimensions, that
$W\ua(\alpha)$ explains. Scenes are split 33/7/8 into training, validation and test
sets, with all views and seeds of a scene in one split. For each model, the block and
timestep are chosen by validation $R^2$ of the difference probe (block 6, $t=200$ for
PixArt-$\alpha$). Intervals are percentile intervals over bootstrap resamples of test
scenes.

\subsection{Single-image readouts and geometric baselines}
\label{sec:readouts}

The plane describes how direction is organised across views; to compare models with
image measurements, we also need decoders that see one image at a time.
\emph{Readouts} predict $\ua(\alpha)$ from the uncentred features $\mathbf h$ of one
image with a linear map and intercept, and the decoded direction is
$\hat\alpha=\operatorname{atan2}(\hat u_2,\hat u_1)$. We report the mean absolute
angular error between $\hat\alpha$ and $\alpha$; a random guess gives $90^\circ$, and
errors above $90^\circ$ mean systematically reversed predictions. Block, timestep and
ridge penalty are chosen by validation error before any held-out path is evaluated,
separately from those of the plane (block 4, $t=50$ for PixArt-$\alpha$). The
\emph{input readout} is fitted in the same way on the model's input embedding, the
tokens before the first block, and shows how much direction information is available
before the model processes the image.
\emph{Geometric baselines} decode $\ua(\alpha)$ from measured image geometry, which
carries most of the perspective information in silhouettes and shadows when the
centre projections are fixed: a linear map on both objects' mask centroids and areas
with derived terms; a linear map that adds each cast shadow's area, offset and second
moments; and a two-layer multilayer perceptron (MLP) on these quantities.

\subsection{Coverage split}
\label{sec:coverage}

The central test is whether a readout keeps its accuracy at camera positions it has not
seen. All probes are fitted on training scenes along the six circles and evaluated on
test scenes along the four ellipses. Because the circles pass through both objects,
they reach the far side of the pair only at short range, so part of each ellipse lies
where no training camera was placed (Fig.~\ref{fig:dataset}c). We express camera
positions in each scene's object-pair frame, with origin at the A--B midpoint and unit
length the A--B separation, and split test views by their distance to the nearest
training camera: \emph{within coverage} if it is below $0.25$, \emph{outside coverage}
if it is at least $0.5$ (\supp{6} reports the intermediate views). We also report the
error on test scenes along the training circles (\emph{familiar paths}), which changes
the scene but not the camera path. Extrapolation here concerns the probe, not the
pretrained model: views outside coverage are new to the readout, but the model's
pretraining data may well contain similar viewpoints. The test asks whether the
model's features hold direction in a form that a map fitted on a limited range of
cameras carries beyond that range.

\subsection{Interventions}
\label{sec:interventions}

Decoding shows that direction is present; interventions test whether later blocks use
the plane and whether it controls the generated image. We edit PixArt-$\alpha$ at one
block, on the tokens inside each object's pooling square. Since each object's tokens
are edited separately, we use the plane fitted to the concatenated features
$[\mathbf a;\mathbf b]$: the upper half of its map, $W_A$, gives the plane's two
directions in object A's features, and the lower half, $W_B$, those in object B's.
\emph{Mean ablation} sets the coordinates of object $o$'s tokens along its two
directions to their training mean; \emph{rotation} adds $W_o[\ua(\alpha+\pi)-\ua(\alpha)]$ to these tokens, which moves
their plane coordinates to those of the opposite direction. Controls apply the same edits
along the top two principal directions of the remaining features, a \emph{variance-matched} subspace (the smallest number of leading
principal directions whose training variance reaches that of the plane, here six),
or a random plane. Effects are read at later blocks by linear decoders of $\alpha$
(trajectory-centred, concatenated features) fitted on unedited training scenes.

To steer generation, we invert held-out renders with denoising diffusion implicit
model (DDIM) inversion~\cite{song2021ddim} and regenerate each one toward the direction
$\alpha_0+\Delta$, where $\alpha_0$ is its true direction and $\Delta$ the requested
rotation. We either \emph{clamp} the plane coordinates at blocks 2--12 to
$\ua(\alpha_0+\Delta)$ at every denoising step, or apply classifier
guidance~\cite{dhariwal2021diffusion} along the latent gradient of
$1-\cos(\hat\alpha-\alpha_0-\Delta)$, with $\hat\alpha$ decoded by a readout at block 6. Random
planes and random readouts serve as controls. The direction of a generated image is
measured by retrieving the nearest render of the same scene (\supp{8}).

\section{Results}
\label{sec:results}

We first compare models with geometric decoders within and outside coverage
(Sec.~\ref{sec:res-coverage}), then ask how the direction code is organised
(Sec.~\ref{sec:res-plane}), which cues the readouts depend on
(Secs.~\ref{sec:res-shape} and~\ref{sec:res-real}), and whether PixArt-$\alpha$ uses
the plane (Sec.~\ref{sec:res-interventions}).

\subsection{Extrapolation beyond the training cameras}
\label{sec:res-coverage}

\begin{table*}[!t]
  \centering
  \caption{\textbf{Eight frozen models} (Tier A spheres, 8 test scenes). Plane: shared-plane
  model on difference features at the validation-selected block and timestep, held-out
  $R^2$. Single image: linear readout at the validation-selected block and timestep;
  mean angular error ($^\circ$) on the training paths (familiar) and on test ellipse views
  within and outside the coverage of training cameras (95\% scene-bootstrap interval),
  and for the same readout at the model input within the coverage.}
  \label{tab:models}
  \small
  \setlength{\tabcolsep}{4pt}
  \begin{tabular}{@{}llcc|ccccc@{}}
    \toprule
    & & \multicolumn{2}{c|}{Plane} & \multicolumn{5}{c}{Single-image error ($^\circ$)}\\
    Model & Training & Block, $t$ & $R^2$ [95\% CI] & Block, $t$ & Familiar & Within & Outside & Input, within\\
    \midrule
    PixArt-$\alpha$ & Diffusion, DiT & 6, 200 & 0.71 [0.69, 0.73] & 4, 50 & 4.6 & 4.5 & 6.5 [6, 7] & 30.6 \\
SD3-medium & Diffusion, MM-DiT & 1, 400 & 0.87 [0.83, 0.90] & 12, 200 & 6.9 & 6.5 & 22.3 [17, 29] & 40.1 \\
DiT-XL/2 & Diffusion, DiT & 6, 100 & 0.68 [0.61, 0.75] & 14, 100 & 4.2 & 4.4 & 8.3 [6, 11] & 20.6 \\
SD 1.5 & Diffusion, UNet & 0, 800 & 0.52 [0.48, 0.56] & 5, 200 & 2.1 & 2.1 & 7.7 [7, 9] & 28.6 \\
DINOv2-L & Self-distillation & 22 & 0.69 [0.65, 0.72] & 16 & 2.6 & 2.6 & 6.8 [6, 8] & 20.0 \\
MAE-L & Masked reconstr. & 4 & 0.53 [0.51, 0.55] & 22 & 2.4 & 2.1 & 6.2 [5, 7] & 10.7 \\
CLIP-L & Image--text & 6 & 0.54 [0.50, 0.58] & 15 & 2.7 & 2.6 & 8.0 [6, 10] & 16.2 \\
SigLIP & Image--text & 8 & 0.41 [0.35, 0.48] & 16 & 3.0 & 3.0 & 7.5 [7, 8] & 19.3 \\
\midrule
Sphere geometry & Linear & -- & -- & -- & 11.1 & 9.3 & 101.6 & -- \\
+ shadows & MLP & -- & -- & -- & 3.6 & 2.4 & 63.4 [46, 78] & -- \\
\bottomrule

  \end{tabular}
\end{table*}

On the training paths, every readout is accurate (Table~\ref{tab:models}, Familiar):
all eight models decode $\alpha$ of held-out scenes to within a few degrees, whereas
the input readout (Sec.~\ref{sec:readouts}) is several times less accurate even
within coverage. The geometric MLP, which decodes $\alpha$ from measured object and
shadow geometry alone, is as accurate as the models, and
remains so on new paths within coverage (Fig.~\ref{fig:teaser}c). On the camera
distribution used for fitting, a learned representation and a decoder of a few image
measurements are therefore indistinguishable.

Outside coverage the two separate (Fig.~\ref{fig:dataset}d). The geometric decoders
fail: the MLP's error rises to $63^\circ$, and the linear decoders exceed $90^\circ$, so
their predictions are reversed. Readouts of seven of the eight models stay within
$6$--$8^\circ$, with scene-bootstrap intervals that do not overlap the MLP's; SD3
extrapolates less well ($22^\circ$). Extrapolation thus holds across training
objectives: diffusion transformers, a diffusion UNet, and self-supervised and
image--text encoders. The PixArt-$\alpha$ plane likewise transfers between small and
large circles, unlike the plane at the input (\supp{6}).

\subsection{A shared, relational plane}
\label{sec:res-plane}

\begin{figure}[!t]
  \centering
  \includegraphics[width=0.82\columnwidth]{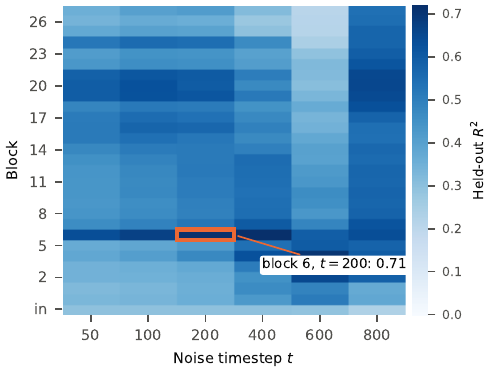}
  \caption{\textbf{Where the plane forms.} Held-out $R^2$ of the shared-plane probe of
  PixArt-$\alpha$ (difference features, test scenes) at the input (in) and at every
  block, for each diffusion timestep $t$. Outlined: the block and timestep selected on
  validation scenes.}
  \label{fig:scan}
\end{figure}

Extrapolation shows that the models encode $\alpha$ beyond the image measurements; we
now ask whether it forms a plane in the hidden space. Every model holds a shared plane
(Table~\ref{tab:models}, Plane), but its strength differs: held-out $R^2$ is highest
in SD3, close to $0.7$ in PixArt-$\alpha$, DINOv2 and DiT-XL/2, lower in CLIP, MAE and
SD~1.5, and lowest in SigLIP. Readout accuracy does not follow plane strength: MAE and
SD~1.5 have the most accurate readouts but weaker planes. A compact, shared code is
thus a property of some representations, and accurate decoding does not require it.

In PixArt-$\alpha$, the plane of the difference features explains $71\%$ of the
variance of held-out centred features. It is relational: a plane fitted to either
object's feature alone explains $5\%$ (held-out trajectories in Fig.~\ref{fig:teaser}b).
The plane is not inherited from the input: at $t=200$, held-out
$R^2$ stays below $0.5$ up to block 5, rises abruptly at block 6 and declines afterwards
(Fig.~\ref{fig:scan}); block 6 is the strongest layer at every $t\le400$, for each
noise seed and prompt (\supp{3}), and also in DiT-XL/2, which receives no text. Block 6's own
update is the best-fitted update of any block, its plane coordinates remain a clean
code for $\alpha$ in every later block, and attention between the two objects' tokens
is near uniform at block 6 and peaks only in blocks 9--21 (\supp{4}): the plane forms
before the objects exchange much information. Shuffling the labels removes the fit,
and with independent noise for every view the plane still explains a third of the
held-out variance (\supp{5}).

\subsection{Shape and appearance}
\label{sec:res-shape}

\begin{figure}[!t]
  \centering
  \includegraphics[width=\columnwidth]{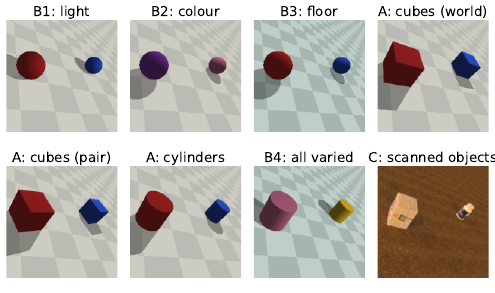}
  \caption{\textbf{Appearance datasets.} The same test scene, camera and direction in
  the Tier A shape datasets, in Tiers B1--B4 and in Tier C (scanned objects,
  path-traced).}
  \label{fig:realism}
\end{figure}

\begin{figure*}[!t]
  \centering
  \includegraphics[width=\textwidth]{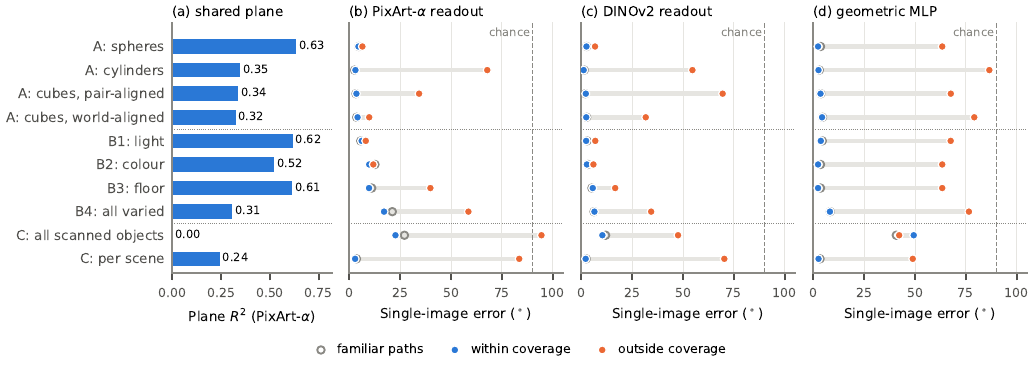}
  \caption{\textbf{Shape and appearance} (Tier A shapes, Tiers B1--B4 and Tier C; 8 test
  scenes; numbers in \supp{9}). (a) Held-out $R^2$ of the dataset's own shared plane in
  PixArt-$\alpha$ (difference features, block 6, $t=50$). (b--d) Single-image angular
  error of the validation-selected readout of PixArt-$\alpha$ and DINOv2 and of the
  geometric MLP, on training paths (grey rings) and on test ellipse views within (blue)
  and outside (orange) the coverage; dashed line: chance. Tier C uses 8-pixel pooling
  windows (other windows and models in \supp{9}). C, per scene: plane and decoders
  fitted within each test scene on its own circles ($R^2$ and familiar error on a
  held-out circle; coverage relative to the scene's own cameras).}
  \label{fig:factors}
\end{figure*}

\begin{table}[!t]
  \centering
  \caption{\textbf{Plane sharing} (PixArt-$\alpha$, block 6, $t=50$). Held-out $R^2$ of
  each dataset's own plane, of the sphere plane on the dataset, and of the dataset's
  plane on spheres; subspace overlap with the sphere plane (mean squared cosine of the
  two principal angles; 1: identical) and rotation of the $(\cos\alpha,\sin\alpha)$
  frame relative to it.}
  \label{tab:sharing}
  \small
  \setlength{\tabcolsep}{3pt}
  \begin{tabular}{@{}lccccc@{}}
    \toprule
    Dataset & Own & Sph.$\rightarrow$set & Set$\rightarrow$sph. & Overlap & Rot.\ ($^\circ$)\\
    \midrule
    A: spheres & 0.63 & 0.63 & 0.63 & 1.00 & 0 \\
A: cylinders & 0.35 & 0.22 & 0.49 & 0.69 & -9 \\
A: cubes (pair) & 0.34 & 0.21 & 0.48 & 0.64 & -5 \\
A: cubes (world) & 0.32 & 0.18 & 0.45 & 0.64 & -11 \\
B1: light & 0.62 & 0.62 & 0.64 & 0.90 & -2 \\
B2: colour & 0.52 & 0.51 & 0.56 & 0.48 & 3 \\
B3: floor & 0.61 & 0.61 & 0.63 & 0.98 & -2 \\
\bottomrule

  \end{tabular}
\end{table}

The sphere results leave open which image cues the readouts use. Changing one factor at
a time (Tier A shapes, Tiers B1--B3) tests this. Within coverage, every readout stays
accurate on every single-factor dataset (Fig.~\ref{fig:factors}, blue). Outside it, PixArt-$\alpha$
extrapolates for spheres, world-aligned cubes, light and colour, but not for
cylinders, pair-aligned cubes or floors; DINOv2 extrapolates for spheres, light and
colour, partly for floors and world-aligned cubes, and not for cylinders and
pair-aligned cubes; the geometric MLP fails throughout. Both models fail on the two
shapes whose visible surfaces are tied to $\alpha$: the top of a cylinder and the faces
of a pair-aligned cube are consistent cues to $\alpha$ among the training cameras, but
they also change with camera elevation, and in our camera layout every view outside
coverage has an elevation not seen near its $\alpha$ during training (\supp{6}). A
readout that relies on these cues is therefore expected to fail outside coverage;
because elevation and coverage coincide in these data, they cannot be separated here.
The faces of world-aligned cubes vary across scenes and cannot serve as such cues.

Strikingly, the plane itself is robust to these factors (Table~\ref{tab:sharing}).
Each factor changes much of the image, and colour changes the very tokens the features
are pooled from, yet a plane fitted on spheres alone carries over: lighting and floor
variation leave it almost unchanged (overlap $0.90$ and $0.98$), and on colour-varied
scenes it explains the features as well as their own plane does. Even replacing the
spheres by cylinders or cubes, whose appearance changes with the view in quite
different ways, preserves much of it: their planes are weaker, but nearly identical to
one another and partly shared with the sphere plane, and the $(\cos\alpha,\sin\alpha)$
frames of all variants agree with that of spheres to within $11^\circ$, at block 6 and
at later blocks (\supp{9}). The model thus places viewing direction along nearly the
same two directions whatever the objects look like. Shape adds direction-dependent
variance on top of this common plane rather than replacing it, so the extrapolation
failures above occur even though the plane is largely unchanged.

\subsection{Varied and real objects}
\label{sec:res-real}

Single factors leave the sphere results largely intact; combining all factors, and
replacing rendered primitives by real objects, tests whether they survive realistic
appearance. When shapes, colours, floor and light all vary (Tier B4,
Fig.~\ref{fig:factors}), the PixArt-$\alpha$ plane weakens to the level of the
non-spherical shapes, and the diffusion transformers lose accuracy even within
coverage (SD3 in \supp{6}). Some loss is expected: unlike the plane probe, the readout
sees uncentred features of one image, and in Tier B4 the pooled squares cover
different surfaces and colours from scene to scene, so appearance mixes with
direction, which a single linear map must separate across only 33 training scenes;
adding appearance variants accordingly halves PixArt-$\alpha$'s error on familiar paths
(\supp{9}). DINOv2, and the geometric MLP,
whose measurements do not depend on colour, keep their accuracy within coverage.
Outside coverage every decoder degrades, DINOv2 least, and more training scenes barely
change PixArt-$\alpha$'s error on ellipses.

With scanned objects (Tier C), no model holds a shared plane (Fig.~\ref{fig:factors};
all models and windows in \supp{9}). Each object pair still has a plane of its own:
fitted within a single scene, the plane explains about a quarter of the variance on a
held-out circle of that scene, and readouts fitted within the scene decode $\alpha$ to
within a few degrees inside coverage (Fig.~\ref{fig:factors}, C per scene). These
planes differ from pair to pair, so no single plane serves all scenes, and, like the
geometric MLP, readouts fitted within one scene do not extrapolate outside coverage. How much of each object the readout sees now
matters for the ViT encoders: widening the pooling window from 8 to 24 pixels lowers the error of DINOv2
and CLIP outside coverage by about $20^\circ$ while changing it within coverage by less
than $5^\circ$. At every window, the SD~1.5 UNet and MAE extrapolate best, and better
than the geometric MLP, whereas both diffusion transformers fail outside coverage at
every window (errors above $90^\circ$), so their failure does not depend on how many
tokens cover each object; SD~1.5 also differs from them in training data and scale.

\subsection{Interventions in PixArt-\texorpdfstring{$\alpha$}{alpha}}
\label{sec:res-interventions}

\begin{table}[!t]
  \centering
  \caption{\textbf{Interventions at block 6}, $t=200$, on 672 held-out views, applied to
  the tokens of each object's pooling square. \emph{Ablate}: the tokens' coordinates in
  a subspace are replaced by their training mean, removing all variation within it.
  \emph{Rotate}: the tokens are shifted by $W_o[\ua(\alpha+180^\circ)-\ua(\alpha)]$, which
  moves their probe-plane coordinates to those of $\alpha+180^\circ$; for the top two PCs
  and the random plane, $W_o$ is replaced by those directions with the probe's column
  norms, giving a shift of the same size.
  PCs: principal components of the features orthogonal to the plane. Dim.: dimension of
  the edited subspace per object; Var.: share (\%) of the object features' training
  variance that lies in it. Error at block: mean error ($^\circ$) of $\alpha$ decoded at a
  later block (27 is the last) by decoders fitted on unedited activations.
  $\|\Delta\epsilon\|$: mean change in the model's noise prediction $\epsilon$; a larger
  value means the edit disturbs the model's output more, whatever its effect on
  direction.}
  \label{tab:ablation}
  \small
  \setlength{\tabcolsep}{2.5pt}
  \resizebox{\columnwidth}{!}{%
  \begin{tabular}{@{}lcc|cccc|c@{}}
    \toprule
    & & & \multicolumn{4}{c|}{Error at block ($^\circ$)} & \\
    Intervention & Dim. & Var. & 7 & 12 & 21 & 27 & $\|\Delta\epsilon\|$\\
    \midrule
    None & -- & -- & 6.7 & 6.2 & 6.1 & 7.1 & 0.00 \\
Ablate probe plane & 2 & 31 & 36.6 & 15.4 & 15.2 & 13.1 & 3.75 \\
Ablate top-2 PCs & 2 & 16 & 8.3 & 6.1 & 6.0 & 8.0 & 8.91 \\
Ablate variance-matched PCs & 6 & 34 & 8.8 & 5.7 & 5.9 & 8.2 & 14.28 \\
Ablate random plane & 2 & 0.2 & 7.0 & 6.3 & 6.0 & 7.0 & 0.68 \\
\midrule
Rotate probe plane $180^\circ$ & 2 & -- & 95.2 & 46.7 & 33.6 & 28.8 & 2.61 \\
Rotate top-2 PCs $180^\circ$ & 2 & -- & 8.9 & 8.8 & 7.5 & 8.7 & 7.14 \\
Rotate random plane $180^\circ$ & 2 & -- & 27.1 & 14.7 & 11.5 & 11.0 & 1.71 \\
\bottomrule

  \end{tabular}}
\end{table}

The analyses so far show that direction is present and organised in a plane; the
interventions of Sec.~\ref{sec:interventions} test whether PixArt-$\alpha$ uses it.
Later blocks still carry $\alpha$, decodable to about $7^\circ$ without any edit
(Table~\ref{tab:ablation}, None), although less of their variance lies in a shared
plane; if this information is derived from the block-6 plane, removing the plane
should degrade it.
Mean-ablating the two-dimensional plane at block 6, which removes about a third of the
variance of each object's features, raises the error of $\alpha$ decoded at block 7 from $7^\circ$ to $37^\circ$, and the error at the final block stays about
twice the unedited level (Table~\ref{tab:ablation}; all blocks in \supp{7}). Ablating
the six-dimensional variance-matched subspace, which removes at least as much variance, leaves
downstream decoding near the unedited level although it changes the noise prediction almost four times as much, and
the top-two and random controls likewise have little effect; the pattern holds at
$t=400$. Rotating the plane by $180^\circ$ removes the original direction without
establishing the new one: at block 7, decoded angles lie about $90^\circ$ from both, and
later blocks recover the original direction (Table~\ref{tab:ablation}), so a direction
written into the plane does not persist. The same rotation applied to the top PCs or
a random plane changes the decoded direction far less, so the effect comes from where
the rotation moves the features, not how far.
Direction
information in later blocks thus passes largely through the plane.

Carrying direction information downstream does not make the plane a control of the
generated scene. On the 8 of 16 renders that DDIM inversion reproduces faithfully,
clamping the plane alters sphere material, shadows and ground texture, but the
generated view moves by at most $7^\circ$ against requested rotations of $90^\circ$ and
$180^\circ$ (\supp{8}). Gradient guidance drives the readout toward the target, yet the
outputs only recolour the spheres or, at high strength, degrade the scene; sizes,
shadows and floor perspective keep the source view. The gradient explains why: the
spheres, $5\%$ of the latent, receive $42\%$ of its mass, a density 14 times that on
the floor and background, whose share is spread too thinly to change the perspective.
This negative result separates two properties that are easily conflated: a variable
can be linearly decodable at a few object tokens, and even carry information
downstream, without being a global control of the scene, whose viewpoint is set by
evidence spread over the floor, shadows and background that this local code does not
reach.

\section{Discussion and Limitations}
\label{sec:discussion}

On camera positions like those used to fit a probe, a learned representation and a
few image measurements decode direction equally well, so accuracy there says little.
Accuracy at camera positions outside that range tells them apart, even though the
model itself may have seen such viewpoints in pretraining. This test has two limits.
First, a shared plane does not guarantee extrapolation: for cylinders and pair-aligned
cubes the plane is partly shared, yet the readouts fail outside coverage, as these
shapes' visible surfaces also change with camera elevation. Second, with real objects
each pair has its own plane, and extrapolation depends on the pooling window and on
the model.

A plane by itself is a weak statement: any feature that varies smoothly and
periodically with $\alpha$ has a first Fourier harmonic in $\cos\alpha$ and $\sin\alpha$,
so a plane within one scene is expected, as we find for each real object pair. What is
nontrivial is that one $W$ serves all scenes, across object sizes, separations, light
directions relative to the pair, noise samples and, in PixArt-$\alpha$, appearance
factors, and carries most of the direction-dependent variance. It does not show that
the model represents an abstract geometric angle: image cues that vary with $\alpha$ in
the same way in every scene would also give a shared plane, and decodability of
$(\cos\alpha,\sin\alpha)$ does not establish that the model computes with $\alpha$.

Rendering is what lets us fix the objects' image positions exactly, so our scenes are
synthetic, with real objects only in Tier C. Natural next steps are to carry
coverage-controlled probing to real photographs and larger scene sets, and to test
the interventions in models beyond PixArt-$\alpha$.

\section{Conclusion}
\label{sec:conclusion}

Vision models trained only on 2D images, whether to generate, reconstruct or match
them with text, encode the viewing direction of a scene in a form that reaches beyond
the cameras a probe was fitted on, where decoders of the visible geometry fail. All
eight store it in the same simple form, one plane shared across scenes in which the
relation between the two objects traces $(\cos\alpha,\sin\alpha)$. In PixArt-$\alpha$
this plane survives changes of appearance and carries direction through the network,
yet writing to it changes the objects rather than the view. Coverage-controlled
probes thus reveal how these models organise viewpoint, and where that organisation
can and cannot be used.

{\small
\bibliographystyle{ieeenat_fullname}
\bibliography{references}
}

\clearpage
\setcounter{section}{0}
\setcounter{figure}{0}
\setcounter{table}{0}
\renewcommand{\thesection}{S\arabic{section}}
\renewcommand{\thetable}{S\arabic{table}}
\renewcommand{\thefigure}{S\arabic{figure}}
\maketitlesupplementary

This supplement provides dataset details and examples (\cref{sec:s-data}),
implementation details (\cref{sec:s-impl}), the full layer--timestep scan and prompt
robustness (\cref{sec:s-scan}), the analysis of where the plane forms
(\cref{sec:s-mech}), controls and single-image decoding at the main analysis point
(\cref{sec:s-controls}), trajectory transfer and coverage for all models (\cref{sec:s-coverage}), the mediation curves (\cref{sec:s-mediation}), the
steering experiments (\cref{sec:s-rotation}), and additional appearance results
(\cref{sec:s-appearance}). Section and table numbers without an ``S'' refer to the
main paper.

\section{Datasets}
\label{sec:s-data}

\begin{figure*}[p]
  \centering
  \includegraphics[width=0.92\textwidth]{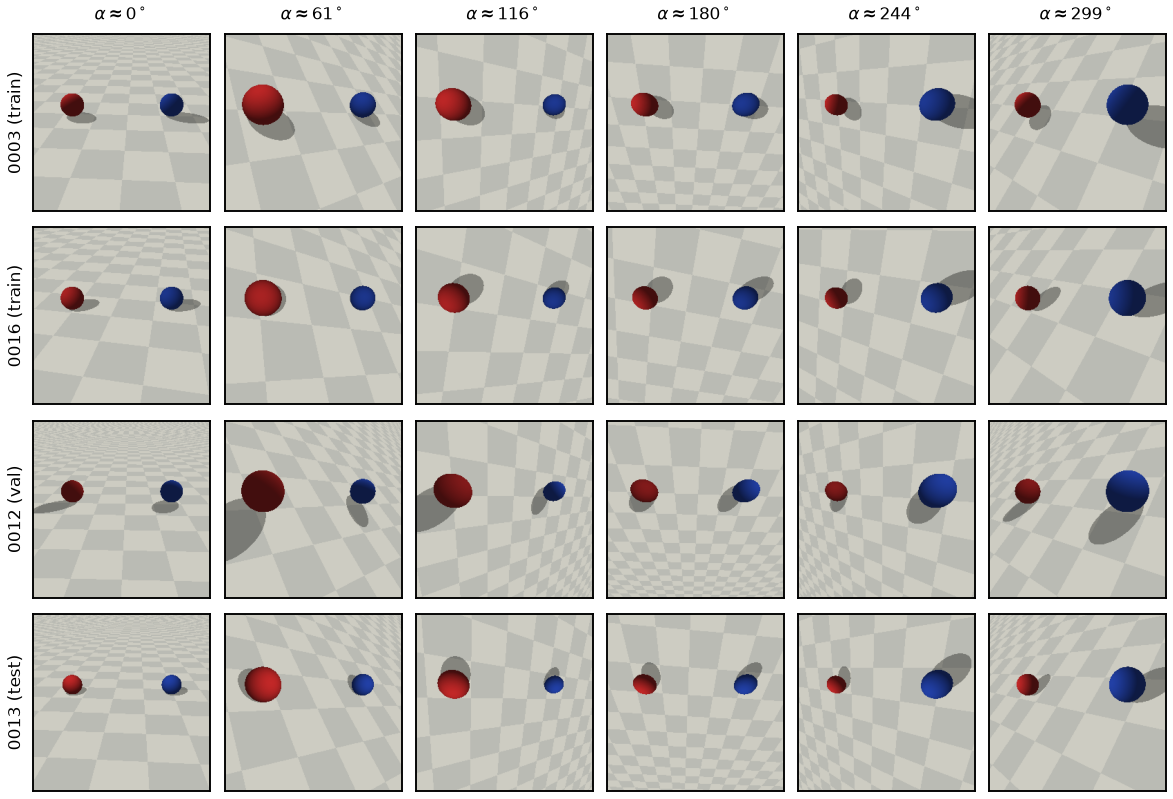}
  \caption{\textbf{Main dataset.} Four scenes (two training, one validation, one test)
  along the source circle at six viewing directions.}
  \label{fig:s-spheres}
\end{figure*}

\paragraph{Camera trajectories.}
All datasets share the camera poses of the main dataset. For every pose, the renderer
solves for focal length and camera orientation so that the centres of A and B project
to the pixels $(56.32, 102.4)$ and $(199.68, 102.4)$ of the $256\times256$ image
(maximum error $1.3\times10^{-12}$ pixels). Of the 48 scenes, twelve share the object
diameter (0.70) and separation (3.0); the other 36 sample the separation in
$[2.4, 3.6]$ and scale the diameter in proportion. Six circles pass through
both object centres in the ground plane (radii 2.0, 2.5, 3.0, 3.32, 4.5 and 5.5) and
four ellipses are centred on the object midpoint with semi-axes and orientations
relative to the A--B chord of $(4.5, 3.0, 0^\circ)$, $(5.5, 2.5, 30^\circ)$,
$(5.0, 3.5, 60^\circ)$ and $(6.0, 2.75, 90^\circ)$. Each trajectory is sampled every
$5^\circ$ of phase; analyses use 14 evenly spaced views per trajectory. Because the
circles pass through the objects, their views sample $\alpha$ non-uniformly (the
trajectory mean of $\ua(\alpha)$ has norm 0.41), whereas the ellipses sample it nearly
uniformly (0.04). \Cref{fig:s-spheres} shows the main dataset.

\begin{figure*}[p]
  \centering
  \includegraphics[width=0.92\textwidth]{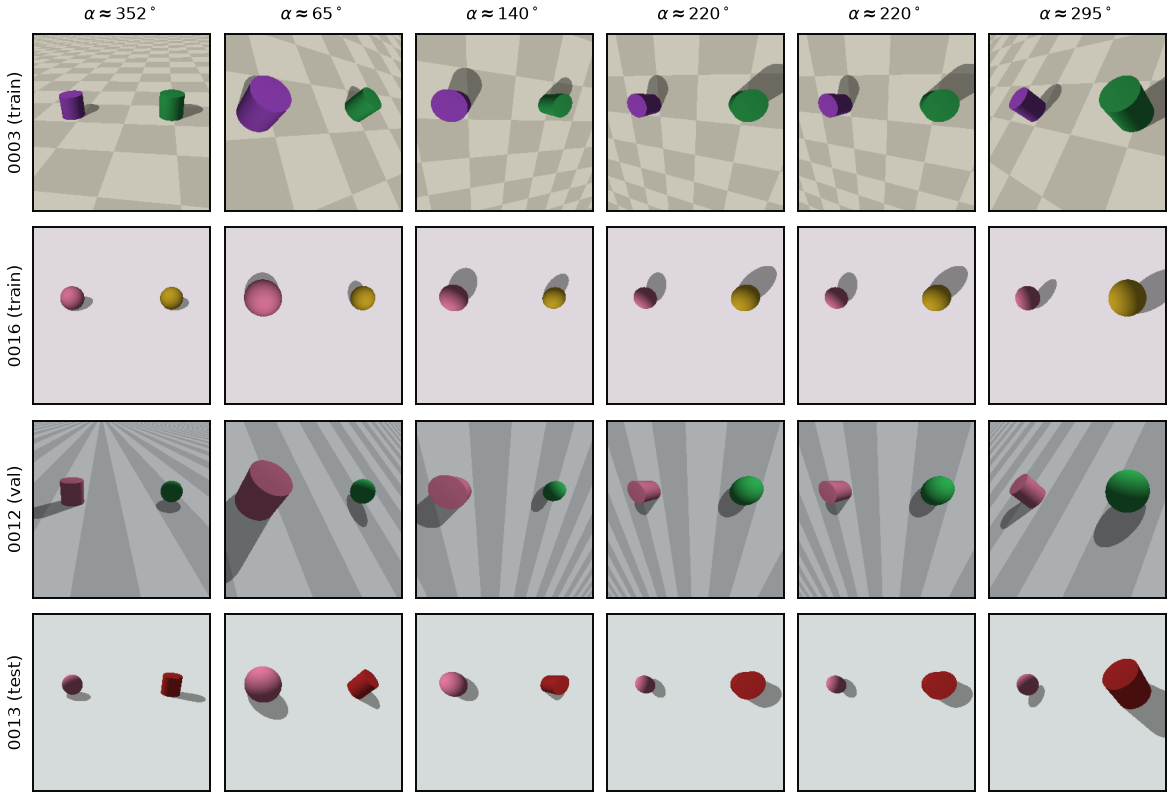}
  \caption{\textbf{Tier B4.} The same scenes and views with shapes, colours, floor and
  light drawn per scene.}
  \label{fig:s-tierAmix}
\end{figure*}

\paragraph{Appearance variants.}
Tier A (cylinders and cubes), Tiers B1--B5 are rendered with the ray caster of
the main dataset, extended with cast shadows for cubes and cylinders and with
optional per-object rotation about the vertical axis. In Tier B5, training scenes use
spheres and cubes and the colours red, blue, orange, cyan, yellow and white;
validation and test scenes always contain a cylinder and use green, purple and pink.
Tier B4 (\cref{fig:s-tierAmix}) draws shapes and all nine colours at random in every
split. Floors are checkered, striped or plain with random tones and tile size; the
point light is placed at a random azimuth around the object midpoint, 5--8 units away
and 5--8 units high. The Tier A shape datasets and Tiers B1--B3 (\cref{fig:s-factors})
keep the main dataset's red and blue spheres, floor and world-fixed light, and change
one property: the shape (Tier A), the light (B1, sampled as in Tier B4), the colours
(B2), or the floor (B3). Cubes and
cylinders have height and width equal to the sphere diameter, so their centres
coincide with the sphere centres and the fixed projections hold exactly. For the
scaling experiment, each of the 33 training scenes of Tier B4 received three
further appearance variants (circular trajectories only), giving 132 training scenes;
validation and test scenes are unchanged, and variants inherit their base scene's
split.

\begin{figure*}[p]
  \centering
  \includegraphics[width=0.92\textwidth]{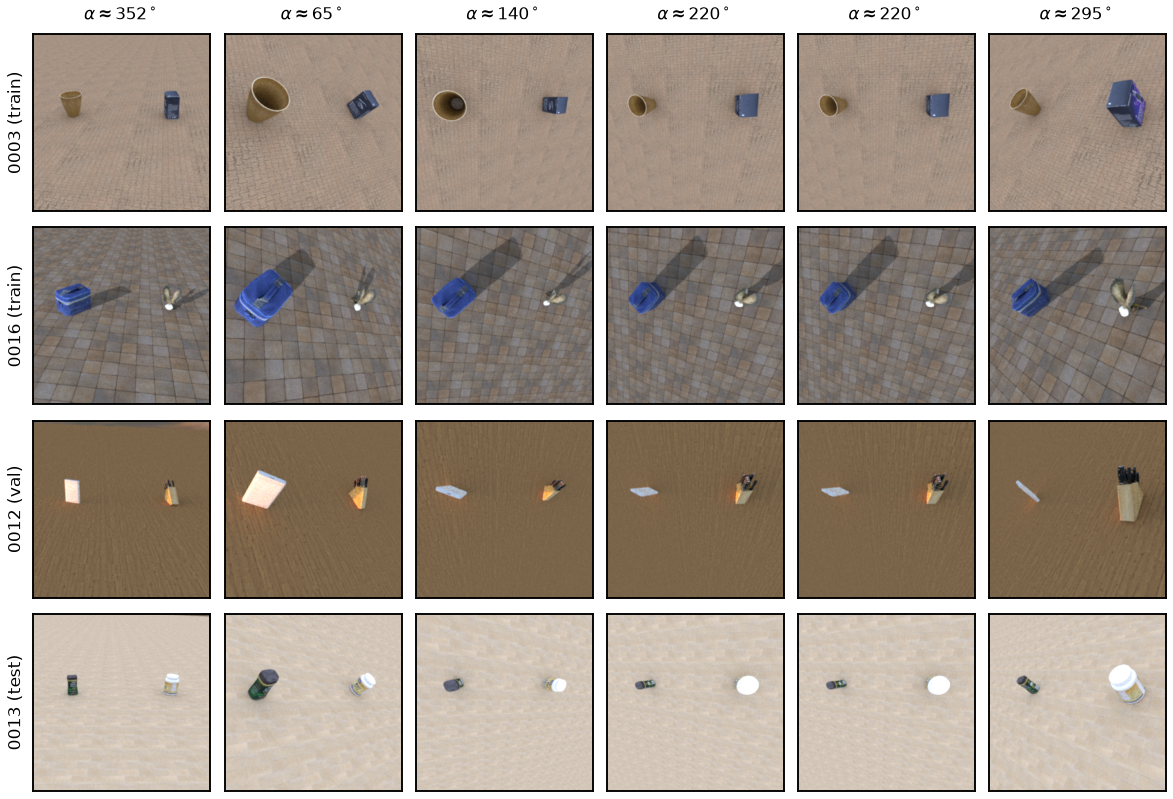}
  \caption{\textbf{Tier C.} The same scenes and views with scanned household objects,
  HDR environment lighting and textured floors.}
  \label{fig:s-tierB}
\end{figure*}

\begin{figure*}[p]
  \centering
  \includegraphics[width=0.78\textwidth]{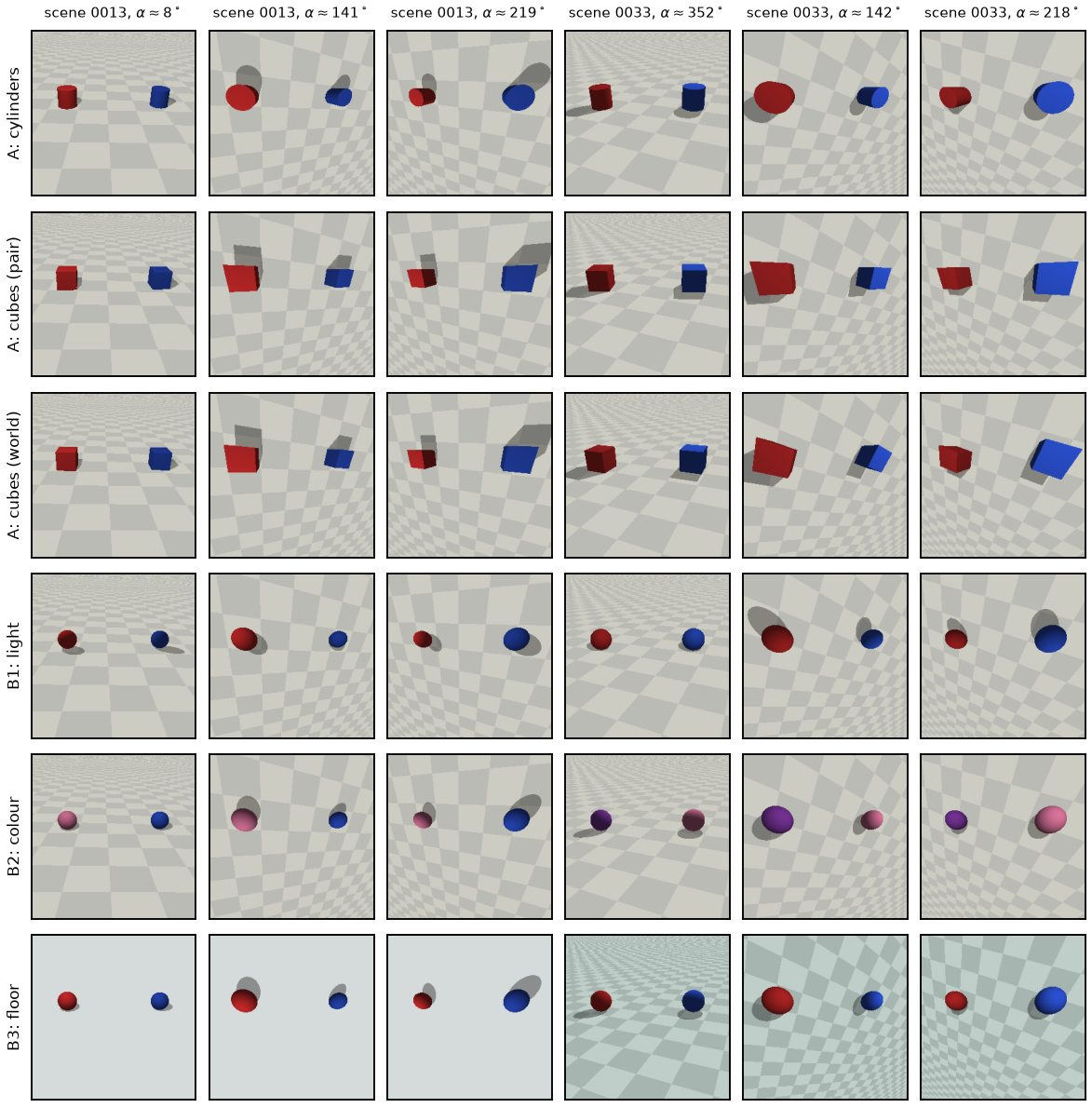}
  \caption{\textbf{Tier A shapes and Tiers B1--B3.} Two test scenes at three viewing
  directions; each row changes one property of the main dataset (Tier A spheres).}
  \label{fig:s-factors}
\end{figure*}

\paragraph{Tier C.}
Each scene contains two objects from Google Scanned Objects, drawn from 500 downloaded
objects whose horizontal-to-vertical extent ratio lies in $[0.5, 1.6]$ (233 usable),
each used in one scene only, rotated by a random yaw and scaled so that its
bounding-box height equals the original sphere diameter and its bounding-box centre
coincides with the original sphere centre. Scenes are lit by one of 12 Poly Haven high-dynamic-range (HDR)
environments (public-domain CC0 licence) with random rotation and placed on one of 12 Poly Haven floor
textures (CC0), path-traced with Mitsuba 3 (48 samples per pixel, maximum depth 6).
Exposure is normalised per scene by the environment's mean luminance. We verified the
camera conversion by rendering the original spheres with Mitsuba and comparing their
masks with the main dataset (intersection over union 0.97--0.99). Object masks come from a shape-index
pass, and shadow masks from the luminance ratio to a second render without the
objects. Three corrections were applied after inspection: one scene whose object was
hollow at its centre (a furniture set) was re-rendered with a replacement object; six
scenes lit by a strongly coloured studio environment (mean RGB 0.60, 0.04, 0.93), whose
renders were clipped to nearly uniform magenta, were re-rendered with neutral
environments; and three views in which an object's centre pixel falls off the object
use the projected centre with zero area as geometric features. All Tier C results
in the main text are computed after these corrections. \Cref{fig:s-tierB} shows
examples.

\FloatBarrier
\section{Implementation details}
\label{sec:s-impl}

\paragraph{Additional models.}
SD~1.5 (UNet in half precision, empty prompt) receives variational-autoencoder (VAE) latents
at $512\times512$ with the same denoising-diffusion (DDPM) noise levels and seeds as DiT-XL/2; its layers are
the outputs of the input convolution ($-1$), the four down blocks (0--3), the middle
block (4) and the four up blocks (5--8), each resampled to a $32\times32$ grid (area
pooling or bilinear interpolation) and zero-padded to 1{,}280 channels, which does not
affect the ridge probes. MAE ViT-L/16 runs at $224\times224$ with masking disabled and an
identity token order; CLIP ViT-L/14 at $336\times336$; SigLIP So400m/14 at $384\times384$; all
use patch tokens only. Pooling squares are defined in pixels of the $256\times256$ render
and resampled to each model's token grid, so a fixed window covers different numbers
of tokens: one token spans 8 pixels of the render for PixArt-$\alpha$, SD3 and
DINOv2 at $448$, 11 for CLIP, 9 for SigLIP, 18 for MAE, and 16 for DINOv2 at 224.

\paragraph{Tier C pooling windows.}
For the comparison at matched footprint (\cref{tab:s-windows}), every layer of each
model is pooled with 8-, 16- and 24-pixel squares at both projected centres from one
forward pass (one noise seed; $t=50$ for diffusion models, the noise level selected on
Tier C for PixArt-$\alpha$ and SD3). For each model and window, the readout layer and
ridge are selected on validation scenes, and the plane is evaluated at the layer with
the best validation $R^2$. At 8 pixels the comparison reproduces the earlier Tier C
results exactly (DINOv2: $10.4^\circ$ within and $47.5^\circ$ outside the coverage). A
DINOv2 run at $224\times224$, whose tokens match MAE's footprint, reaches $14.4^\circ$ and
$32.2^\circ$ with the 8-pixel window, between DINOv2 at 448 and MAE
(\cref{tab:s-coverage}).

\paragraph{Feature extraction.}
Images are resized to each model's resolution (PixArt-$\alpha$ and SD3: 512, DiT-XL/2:
256, DINOv2: 448). Diffusion latents use the VAE posterior mode. PixArt-$\alpha$ and
DiT-XL/2 are noised with their DDPM forward process. For SD3, which is trained with
rectified flow, $x_t=(1-\sigma)x_0+\sigma\epsilon$, we choose $\sigma$ so that
$(1-\sigma)/\sigma=\sqrt{\bar\alpha_t/(1-\bar\alpha_t)}$ under the linear DDPM schedule,
giving $\sigma=0.15, 0.42, 0.67, 0.96$ for $t=50, 200, 400, 800$. We capture the
image-token stream after every block (for SD3, the image stream of each joint block)
and the patch-embedded input. Noise is shared by all views of a scene (matched noise)
and depends only on the scene and seed, so all datasets with the same scenes see the
same noise. SD3 uses its two CLIP text encoders without the T5 text encoder and, like SD~1.5, an empty prompt;
PixArt-$\alpha$ uses ``a scene containing two geometric objects''.

\paragraph{Probes and selection.}
The shared-plane model is fitted by ridge regression ($\lambda=10^{-3}$) on
trajectory-centred features; its analysis point is chosen by validation $R^2$.
Single-image readouts are ridge decoders with an intercept on uncentred concatenated
features. The layer and timestep are chosen by validation error on training
trajectories with $\lambda\in\{10^{-1},10,10^{3},10^{5}\}$; at the chosen point, $\lambda$ is
re-selected on the same validation scenes from $\{10^{-3},10^{-1},10,10^{3},10^{5}\}$. Geometric decoders use
standardised features and $\lambda\in\{10^{-3},10^{-1},10,10^{3}\}$; the MLP has two
hidden layers of width 64 or 256 and weight decay in $\{10^{-4},\dots,10^{-1}\}$, with
early stopping, chosen on the same validation scenes. All confidence intervals are
percentile intervals over 1{,}000 bootstrap resamples of test scenes.

\FloatBarrier
\section{Layer--timestep scan and prompts}
\label{sec:s-scan}

Fig.~3 of the main text shows the held-out $R^2$ of the shared-plane model for
PixArt-$\alpha$ at every layer and noise level. \Cref{tab:s-prompts} repeats the scan at $t=200$ with
two alternative prompts; the profile, including the block-6 peak, is unchanged.

\begin{table}[!htbp]
  \centering
  \caption{\textbf{Prompt robustness.} Held-out $R^2$ at $t=200$ for the main prompt
  (``a scene containing two geometric objects'', three seeds), an empty prompt, and
  ``a red sphere and a blue sphere on a checkered floor'' (one seed each).}
  \label{tab:s-prompts}
  \small
  \begin{tabular}{@{}lccc@{}}
    \toprule
    Block & Main & Empty & Descriptive\\
    \midrule
    input & 0.297 & 0.297 & 0.297 \\
3 & 0.374 & 0.375 & 0.373 \\
4 & 0.472 & 0.472 & 0.475 \\
5 & 0.416 & 0.420 & 0.417 \\
6 & 0.713 & 0.710 & 0.718 \\
7 & 0.524 & 0.514 & 0.526 \\
8 & 0.493 & 0.476 & 0.489 \\
12 & 0.514 & 0.494 & 0.523 \\
19 & 0.614 & 0.605 & 0.619 \\
20 & 0.597 & 0.584 & 0.593 \\
21 & 0.599 & 0.589 & 0.603 \\
27 & 0.414 & 0.406 & 0.406 \\
\bottomrule

  \end{tabular}
\end{table}

\FloatBarrier
\section{Where the plane forms}
\label{sec:s-mech}

\begin{figure}[!htbp]
  \centering
  \includegraphics[width=\columnwidth]{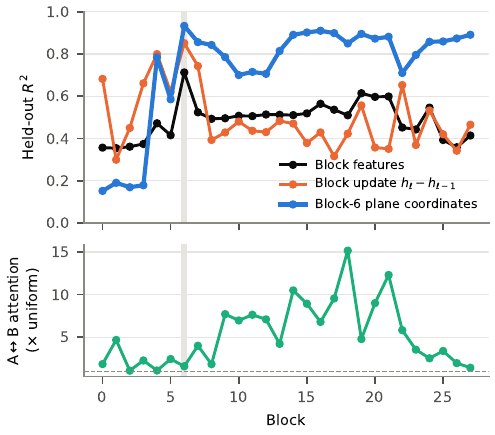}
  \caption{\textbf{Where the plane forms} ($t=200$, test scenes). Grey bar: block 6.}
  \label{fig:s-mech}
\end{figure}

\Cref{fig:s-mech} reports, for PixArt-$\alpha$ at $t=200$, the held-out $R^2$ of the
shared-plane model for each block's output, for each block's update
$h_\ell-h_{\ell-1}$, and for the two coordinates of the block-6 plane read at every
block, together with the attention mass between the support tokens of A and B
relative to uniform attention, computed from each block's own query and key
projections on 96 test views. Block 6 writes the best-fitted update ($R^2=0.85$), the
block-6 coordinates remain a clean code thereafter, and cross-object attention is low
at block 6 and peaks in blocks 9--21.

\FloatBarrier
\section{Controls and single-image decoding}
\label{sec:s-controls}

\begin{table}[!htbp]
  \centering
  \caption{\textbf{Shared-plane probe of PixArt-$\alpha$ on held-out scenes} (block 6 and
  input, $t=200$; 8 test scenes). Plane: fraction of centred feature energy in the
  learned plane (\%). Error: mean angular error of the trajectory-centred decoder
  ($^\circ$).}
  \label{tab:s-main}
  \small
  \setlength{\tabcolsep}{3pt}
  \resizebox{\columnwidth}{!}{%
  \begin{tabular}{@{}llcccc@{}}
    \toprule
    Features & Rep. & $R^2$ & 95\% CI & Plane & Error\\
    \midrule
    Input & Concat. & 0.178 & [0.130, 0.223] & 52.8 & 17.8 \\
Input & Diff. & 0.297 & [0.246, 0.346] & 68.9 & 24.0 \\
Block 6 & Concat. & 0.472 & [0.455, 0.490] & 55.7 & 5.9 \\
Block 6 & Diff. & 0.713 & [0.693, 0.733] & 76.4 & 6.8 \\
\bottomrule

  \end{tabular}}
\end{table}

\Cref{tab:s-main} gives the shared-plane probe of PixArt-$\alpha$ at the
validation-selected point for both representations. Concatenated features fit the
plane less completely ($R^2=0.472$) but give the most accurate trajectory-centred
decoding ($5.9^\circ$). Per-scene $R^2$ is 0.735--0.762 for the three fixed-geometry test
scenes and 0.661--0.717 for the five with unseen object size and spacing. At block 6,
$b+W\ua(\alpha)$ explains 14\% of the total held-out variance of uncentred features, most
of which reflects differences between scenes.

\begin{table}[!htbp]
  \centering
  \caption{\textbf{Controls} at block 6, $t=200$ (held-out $R^2$).}
  \label{tab:s-controls}
  \small
  \begin{tabular}{@{}lrr@{}}
    \toprule
    Condition & Concat. & Difference\\
    \midrule
    Matched noise (main) & 0.472 & 0.713 \\
Independent noise per view & 0.205 & 0.339 \\
Shuffled labels (mean of 100) & 0.000 & 0.000 \\
Shuffle $p$-value & 0.0099 & 0.0099 \\
\bottomrule

  \end{tabular}
\end{table}

\begin{table}[!htbp]
  \centering
  \caption{\textbf{Single-image decoding} on held-out scenes along the training
  trajectories (PixArt-$\alpha$; mean angular error in degrees, 95\% scene-bootstrap
  interval).}
  \label{tab:s-single}
  \small
  \begin{tabular}{@{}lcc@{}}
    \toprule
    Decoder & Error & 95\% CI\\
    \midrule
    DiT block 4, $t=50$ (linear) & 4.6 & [3.9, 5.5] \\
DiT input (linear) & 28.9 & [25.4, 31.8] \\
Spheres (linear) & 11.1 & [9.0, 13.2] \\
Spheres + shadows (linear) & 10.4 & [8.7, 12.5] \\
Spheres + shadows (MLP) & 3.6 & [2.7, 4.7] \\
\bottomrule

  \end{tabular}
\end{table}

\Cref{tab:s-controls} lists the label-shuffle and independent-noise controls of
Sec.~4.2, and \cref{tab:s-single} the single-image decoders on familiar trajectories
summarised in Fig.~1c.

\FloatBarrier
\section{Trajectory transfer and coverage}
\label{sec:s-coverage}

\paragraph{Transfer across trajectory families.}
\Cref{tab:transfer} lists the transfer results summarised in Sec.~4.1: the plane and all
decoders are fitted on training scenes along one family of trajectories and evaluated
on test scenes along another. Errors on ellipses are averaged over all test ellipse
views, irrespective of coverage, and over three noise seeds.

\begin{table*}[t]
  \centering
  \caption{\textbf{Transfer to held-out camera trajectories} (test scenes; plane at
  block 6, $t=200$; single-image readout at block 4, $t=50$).
  The plane and all decoders are fitted on training scenes and training trajectories
  only. $R^2$: trajectory-centred shared-plane model (difference features, block 6) on
  training (seen) and held-out trajectories. Error: single-image angular decoding
  error on held-out trajectories ($^\circ$; best in bold) for linear decoders on DiT
  features at the validation-selected readout (block 4, $t=50$) and at the input, and
  for decoders on measured sphere and shadow geometry.}
  \label{tab:transfer}
  \small
  \begin{tabular}{@{}lcc|ccccc@{}}
    \toprule
    & \multicolumn{2}{c|}{Plane $R^2$} & \multicolumn{5}{c}{Single-image error on held-out trajectories ($^\circ$)}\\
    Protocol & Seen & Held-out [95\% CI] & Readout & Input & Spheres & +Shadows & +Shadows (MLP)\\
    \midrule
    Small $\rightarrow$ large radii & 0.74 & 0.63 [0.60, 0.66] & 8.4 & 31.8 & 15.3 & 15.0 & \textbf{5.6} \\
Large $\rightarrow$ small radii & 0.69 & 0.71 [0.67, 0.75] & \textbf{8.9} & 35.5 & 20.4 & 18.5 & 13.5 \\
Circles $\rightarrow$ ellipses & 0.71 & 0.56 [0.49, 0.61] & \textbf{5.6} & 62.2 & 52.7 & 48.9 & 30.8 \\
\bottomrule

  \end{tabular}
\end{table*}

\paragraph{Coverage bins.}
\Cref{tab:s-coverage} lists single-image errors on test ellipse views in all three
coverage bins, for every model and dataset in the main text; the intermediate bin
($0.25$--$0.5$ object separations, 45 of 448 views) is omitted there. Bins depend only on
the camera poses and are therefore identical for all datasets. Readouts are trained on
all noise seeds of the circle caches (three for diffusion models); ellipse caches have
one seed, except for SD~1.5. Intervals in the main text resample the eight test
scenes (2{,}000 samples). Every test view outside the coverage has a camera elevation
that no training view within $10^\circ$ of its $\alpha$ has, so coverage and elevation
cannot be separated in these data. Errors above $90^\circ$ outside the
coverage, which occur mainly for readouts at the model input, indicate predictions
that are systematically reversed rather than random (chance level is $90^\circ$).

\begin{table*}[t]
  \centering
  \caption{\textbf{Single-image error by coverage} (test ellipse views, 8 test scenes,
  readouts trained on all noise seeds; degrees). Within: distance to the nearest training camera $<0.25$
  object separations (200 views); near: $0.25$--$0.5$ (45); outside: $\ge0.5$ (203).
  Readout: validation-selected block and timestep; input: the same procedure at the
  model input. The geometric decoders use measured object and shadow geometry. Tier C
  rows use the 8-pixel window; \cref{tab:s-windows} compares windows.}
  \label{tab:s-coverage}
  \footnotesize
  \begin{tabular}{@{}llc|ccc|ccc@{}}
    \toprule
    & & & \multicolumn{3}{c|}{Readout} & \multicolumn{3}{c}{Input}\\
    Data & Model & Block, $t$ & Within & Near & Outside & Within & Near & Outside\\
    \midrule
    A: spheres & PixArt-$\alpha$ & 4, 50 & 4.5 & 7.0 & 6.5 & 30.6 & 67.6 & 92.1 \\
 & SD3-medium & 12, 200 & 6.5 & 16.1 & 22.3 & 40.1 & 61.3 & 81.9 \\
 & DiT-XL/2 & 14, 100 & 4.4 & 6.8 & 8.3 & 20.6 & 48.1 & 73.0 \\
 & SD 1.5 & 5, 200 & 2.1 & 3.9 & 7.7 & 28.6 & 63.4 & 78.7 \\
 & DINOv2-L & 16 & 2.6 & 5.5 & 6.8 & 20.0 & 52.0 & 96.2 \\
 & MAE-L & 22 & 2.1 & 2.7 & 6.2 & 10.7 & 22.8 & 77.8 \\
 & CLIP-L & 15 & 2.6 & 4.3 & 8.0 & 16.2 & 45.7 & 97.8 \\
 & SigLIP & 16 & 3.0 & 3.8 & 7.5 & 19.3 & 55.4 & 79.8 \\
 & Geometry, linear & -- & 9.3 & 24.7 & 101.6 & & & \\
 & Geometry, MLP & -- & 2.4 & 10.1 & 63.4 & & & \\
\midrule
A: cylinders & PixArt-$\alpha$ & 6, 50 & 3.0 & 19.5 & 67.8 & 24.6 & 47.8 & 109.1 \\
 & DINOv2-L & 16 & 1.2 & 5.6 & 54.6 & 12.4 & 35.7 & 98.3 \\
 & Geometry, linear & -- & 7.1 & 22.3 & 104.4 & & & \\
 & Geometry, MLP & -- & 2.6 & 13.1 & 86.5 & & & \\
\midrule
A: cubes (pair) & PixArt-$\alpha$ & 5, 50 & 3.5 & 11.1 & 34.3 & 33.9 & 60.7 & 113.5 \\
 & DINOv2-L & 18 & 2.3 & 10.2 & 69.5 & 19.8 & 59.3 & 97.1 \\
 & Geometry, linear & -- & 11.0 & 32.1 & 98.0 & & & \\
 & Geometry, MLP & -- & 3.6 & 14.0 & 67.6 & & & \\
\midrule
A: cubes (world) & PixArt-$\alpha$ & 5, 50 & 4.1 & 7.2 & 9.8 & 34.5 & 61.4 & 110.5 \\
 & DINOv2-L & 18 & 2.5 & 7.0 & 31.7 & 18.5 & 43.9 & 90.6 \\
 & Geometry, linear & -- & 11.3 & 26.2 & 94.2 & & & \\
 & Geometry, MLP & -- & 4.3 & 13.1 & 79.1 & & & \\
\midrule
B1: light & PixArt-$\alpha$ & 5, 50 & 6.1 & 6.0 & 8.1 & 33.1 & 61.6 & 102.0 \\
 & DINOv2-L & 23 & 2.4 & 2.3 & 6.9 & 25.4 & 42.5 & 98.2 \\
 & Geometry, linear & -- & 9.3 & 24.7 & 101.6 & & & \\
 & Geometry, MLP & -- & 3.7 & 8.7 & 67.5 & & & \\
\midrule
B2: colour & PixArt-$\alpha$ & 5, 50 & 9.8 & 10.9 & 11.8 & 33.4 & 87.5 & 119.6 \\
 & DINOv2-L & 19 & 2.8 & 5.2 & 6.0 & 24.9 & 44.2 & 75.4 \\
 & Geometry, linear & -- & 9.3 & 24.7 & 101.6 & & & \\
 & Geometry, MLP & -- & 2.4 & 10.1 & 63.4 & & & \\
\midrule
B3: floor & PixArt-$\alpha$ & 13, 50 & 9.7 & 13.9 & 39.8 & 32.7 & 85.2 & 113.2 \\
 & DINOv2-L & 20 & 5.6 & 5.9 & 16.7 & 20.9 & 48.4 & 92.9 \\
 & Geometry, linear & -- & 9.3 & 24.7 & 101.6 & & & \\
 & Geometry, MLP & -- & 2.4 & 10.1 & 63.4 & & & \\
\midrule
B4: all varied & PixArt-$\alpha$ & 10, 50 & 17.1 & 33.8 & 58.5 & 41.5 & 70.3 & 112.6 \\
 & SD3-medium & 13, 200 & 17.7 & 45.4 & 87.0 & 40.3 & 67.6 & 90.3 \\
 & DINOv2-L & 23 & 6.5 & 17.8 & 34.3 & 38.4 & 54.7 & 107.7 \\
 & Geometry, linear & -- & 17.2 & 36.0 & 105.3 & & & \\
 & Geometry, MLP & -- & 8.2 & 15.6 & 76.5 & & & \\
\midrule
C: scanned & PixArt-$\alpha$ & 13, 50 & 22.7 & 39.3 & 94.4 & 43.5 & 83.7 & 142.6 \\
 & SD3-medium & 12, 50 & 21.2 & 30.4 & 107.7 & 42.7 & 84.9 & 143.7 \\
 & SD 1.5 & 5, 50 & 6.8 & 6.7 & 14.4 & 42.7 & 82.8 & 142.7 \\
 & DINOv2-L & 19 & 10.4 & 15.9 & 47.5 & 42.7 & 84.2 & 144.1 \\
 & DINOv2-L (224) & 17 & 14.4 & 17.2 & 32.2 & 42.6 & 84.2 & 144.2 \\
 & MAE-L & 21 & 11.9 & 18.3 & 21.8 & 34.7 & 82.4 & 137.4 \\
 & CLIP-L & 14 & 24.5 & 29.2 & 56.1 & 42.5 & 84.7 & 144.3 \\
 & SigLIP & 11 & 21.4 & 21.8 & 46.5 & 43.4 & 85.8 & 143.3 \\
 & Geometry, linear & -- & 35.2 & 55.5 & 117.0 & & & \\
 & Geometry, MLP & -- & 49.4 & 41.3 & 42.2 & & & \\
\bottomrule

  \end{tabular}
\end{table*}

\FloatBarrier
\section{Mediation curves}
\label{sec:s-mediation}

\begin{figure}[!htbp]
  \centering
  \includegraphics[width=0.95\columnwidth]{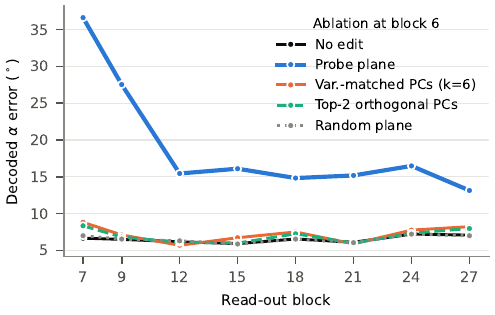}
  \caption{\textbf{Mediation.} Error of $\alpha$ decoded at each later block after mean
  ablation at block 6 ($t=200$, 672 test views). Only ablation of the probe plane
  affects downstream direction information.}
  \label{fig:s-mediation}
\end{figure}

\Cref{fig:s-mediation} plots, for every block after the intervention, the error of
the clean-trained decoders after mean-ablating the probe plane, the variance-matched
principal subspace ($k=6$), the top two principal directions, or a random plane
(Table 3 lists selected blocks).

\FloatBarrier
\section{Steering generation}
\label{sec:s-rotation}

\paragraph{Clamping: procedure.}
A held-out render is inverted to $t=800$ with 50 DDIM steps using the conditional
branch only and regenerated with the same schedule. At every step and at blocks 2--12,
let $\bar h$ be the pooled concatenated object features, $\mu_{\text{traj}}$ the
trajectory mean at that block and the nearest cached timestep, and $W$ the
concatenation probe. With current plane coordinates $c=W^{+}(\bar h-\mu_{\text{traj}})$,
we add $W[\ua(\alpha_0+\Delta)-c]$ to the support tokens of each object (the A and B
halves respectively). Because the coordinates are set rather than incremented, edits
at successive blocks do not compound. The control applies the same procedure to a
random plane with the probe's per-object column norms. Outputs are compared with the
true render at $\alpha_0+\Delta$ by nearest-neighbour retrieval among all
circular-trajectory renders of the same scene ($64\times64$ RGB) and by the ratio of
segmented sphere areas.

\begin{figure*}[t]
  \centering
  \includegraphics[width=0.95\textwidth]{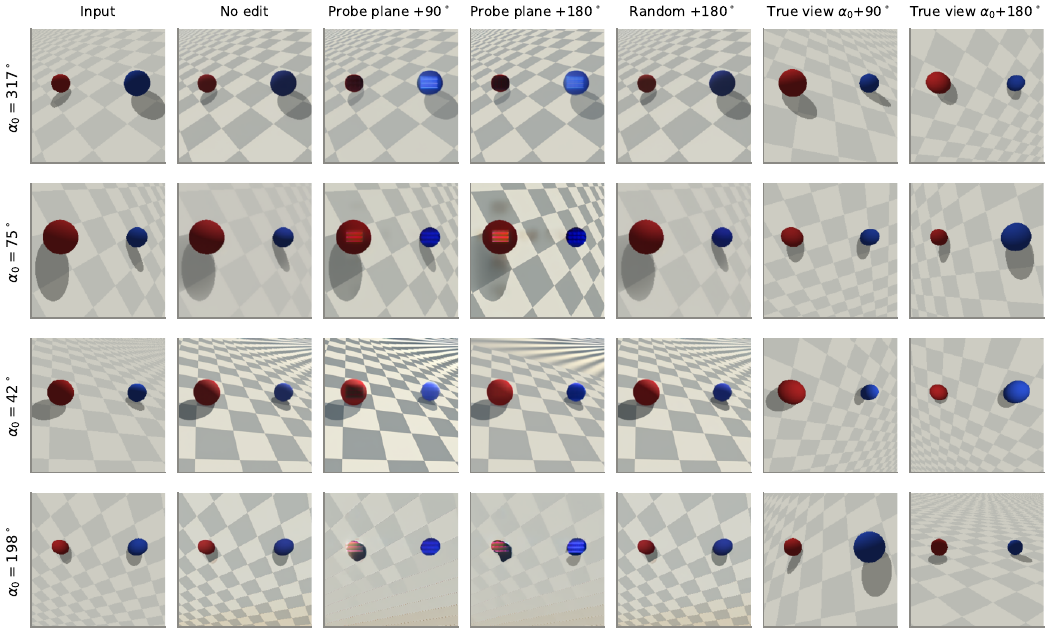}
  \caption{\textbf{Clamping during generation.} Faithfully inverted sources. Clamping
  the probe plane alters material, shadows and ground texture but does not produce
  the viewpoint of the true renders (right); a random plane driven by the same
  procedure changes the image little.}
  \label{fig:s-rotation}
\end{figure*}

\paragraph{Clamping: results.}
Of 16 sources (two per test scene), DDIM inversion reproduces 8 faithfully (pixel root-mean-square error, RMSE,
$\le0.07$ at $128\times128$); the other 8 already depart from the input without
editing (RMSE $\ge0.16$) and are excluded. \Cref{tab:s-rotation} gives the statistics
on the faithful sources and \cref{fig:s-rotation} examples.

\begin{table}[!htbp]
  \centering
  \caption{\textbf{Clamping during generation} (8 faithful sources). Retrieved shift:
  angular difference between the renders retrieved for edited and unedited outputs
  (degrees; mean and maximum). Toward: outputs whose change in
  $\log(\text{area}_B/\text{area}_A)$ has the sign of the true change (\%). Median change
  of this log ratio for edited outputs and for the true renders.}
  \label{tab:s-rotation}
  \small
  \setlength{\tabcolsep}{2.5pt}
  \resizebox{\columnwidth}{!}{%
  \begin{tabular}{@{}lcccccc@{}}
    \toprule
    & & \multicolumn{2}{c}{Retrieved shift} & & \multicolumn{2}{c}{Median $\Delta$ log ratio}\\
    Plane & $\Delta$ & Mean & Max & Toward & Output & True\\
    \midrule
    Probe plane & $+90^\circ$ & 1.9 & 7.2 & 38 & 0.07 & $-$0.95 \\
Probe plane & $+180^\circ$ & 1.0 & 4.5 & 50 & 0.04 & $-$1.14 \\
Random plane & $+90^\circ$ & 0.0 & 0.0 & 12 & 0.00 & $-$0.95 \\
Random plane & $+180^\circ$ & 0.0 & 0.0 & 12 & 0.00 & $-$1.14 \\
\bottomrule

  \end{tabular}}
\end{table}

\paragraph{Gradient guidance: procedure.}
Gradient guidance uses the block-6 readout as a judge and changes the image until the
judge reports the target direction, as in classifier
guidance (Sec.~3.7 of the main text). Unlike clamping, which overwrites hidden
features, it changes the latent image itself. A held-out render is DDIM-inverted to
$t=600$ or $t=800$ (50-step schedule) and regenerated with the same schedule. At each
denoising step: (i) the current latent runs through blocks 0--6, the object tokens are
pooled as in the activation caches, and a single-image readout fitted on training
scenes gives the direction $\hat\alpha$; (ii) the mismatch $L=1-\cos(\hat\alpha-\alpha^*)$
with the target $\alpha^*=\alpha_0+90^\circ$ is differentiated with respect to the
latent; (iii) this gradient, normalised and scaled by a strength $s$, is added to the
model's noise prediction before the denoising step. The control uses a random readout
of the same size.

\begin{figure*}[t]
  \centering
  \includegraphics[width=0.85\textwidth]{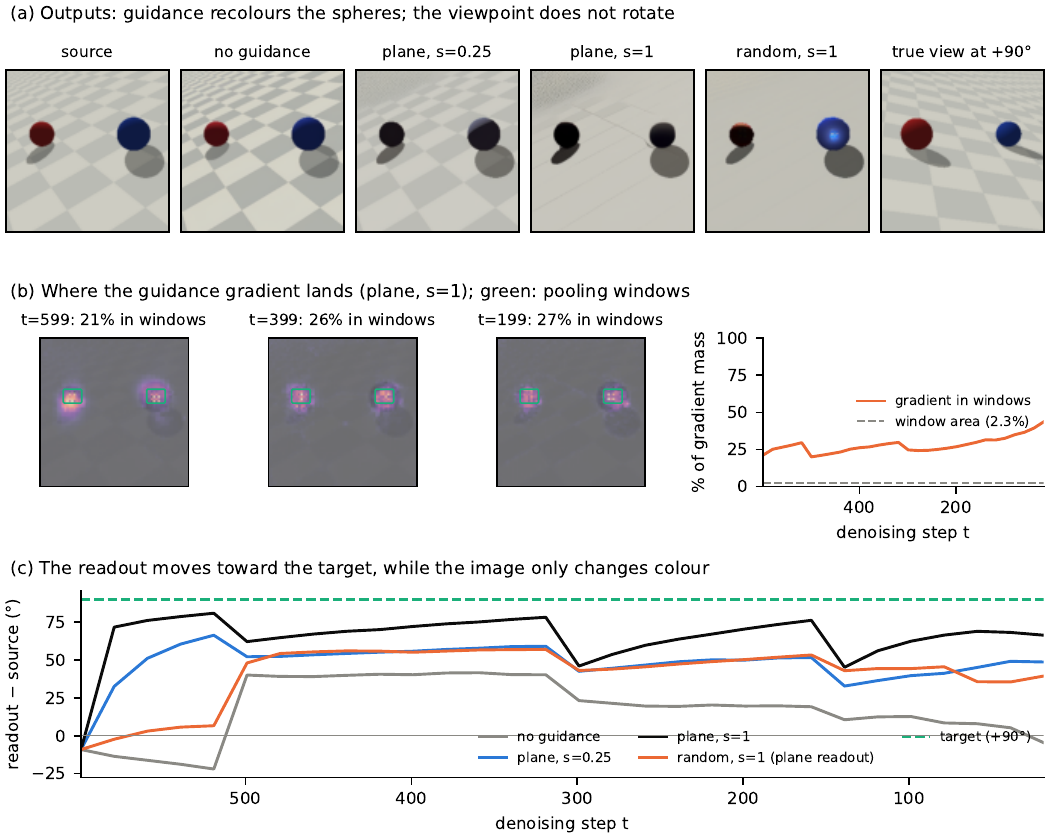}
  \caption{\textbf{Gradient guidance toward $+90^\circ$} (one source, inverted to
  $t=600$). (a) Source, output without guidance, outputs with plane guidance at
  strengths $s=0.25$ and $1$ and with a random readout, and the true view at the target.
  (b) Magnitude of the guidance gradient over the latent at three denoising steps
  (green: pooling windows), and the share of gradient mass inside the windows. (c)
  Readout minus source angle over the denoising steps; jumps occur where the readout
  switches to the next cached timestep, and the unguided readout also drifts at high
  noise.}
  \label{fig:s-guidance}
\end{figure*}

\paragraph{Gradient guidance: results.}
Without guidance, the inverted sources are reproduced faithfully. With guidance, the
readout moves toward the target and can reach it (within $10^\circ$ at $s=2$ from
$t=800$), but the image keeps its viewpoint: only the spheres change colour, and at
larger $s$ the floor and lighting degrade (\cref{fig:s-guidance}a). No output shows
the size ratio, shadows or floor perspective of the target view. The distribution of
the gradient shows why (\cref{fig:s-guidance}b). Because the readout looks only at the
object tokens, the gradient lands mostly on the spheres: averaged over four sources
and all steps ($s=1$), the spheres (5.0\% of the latent) receive 42\% of the gradient
mass, whereas the floor and background (90.5\% of the latent), which carry the
perspective, receive 52\%, 14 times less per unit area. The cheapest way to change the
readout's answer is to repaint the objects, not to move the camera. This negative
result is informative: a representation that is linearly decodable from, and editable
at, a few object tokens need not be a variable through which the global scene can be
controlled.

\FloatBarrier
\section{Additional appearance results}
\label{sec:s-appearance}

\paragraph{All runs.}
\Cref{tab:s-tiers} lists every appearance run, including Tier B5 with held-out shapes
and colours and both noise levels for the diffusion models on Tier B4; the main
paper reports, per model and dataset, the run with the lower validation error.

\begin{table*}[t]
  \centering
  \caption{\textbf{All appearance runs} (8 test scenes). Readout layer and timestep
  chosen on validation scenes (validation error shown); single-image angular error
  (degrees) on familiar and elliptical trajectories for the readout, the geometric MLP,
  and the readout applied at the model input.}
  \label{tab:s-tiers}
  \small
  \begin{tabular}{@{}llcc|cc|cc|c@{}}
    \toprule
    & & & & \multicolumn{2}{c|}{Readout} & \multicolumn{2}{c|}{Geometry MLP} & Input\\
    Data & Model & Block, $t$ & Val. & Familiar & Ellipses & Familiar & Ellipses & Ellipses\\
    \midrule
    B5 & PixArt-$\alpha$ & 23, 200 & 16.0 & 16.6 & 47.7 & 9.4 & 42.7 & 80.8 \\
\midrule
B4 & PixArt-$\alpha$ & 14, 200 & 13.6 & 21.3 & 35.9 & 8.3 & 39.9 & 85.7 \\
 & PixArt-$\alpha$ & 10, 50 & 11.7 & 21.2 & 37.5 & 8.3 & 39.9 & 76.6 \\
 & SD3-medium & 13, 200 & 11.7 & 20.0 & 51.9 & 8.3 & 39.9 & 65.7 \\
 & SD3-medium & 16, 50 & 13.9 & 27.8 & 51.3 & 8.3 & 39.9 & 58.2 \\
 & DINOv2-L & 23 & 7.1 & 6.3 & 20.2 & 8.3 & 39.9 & 71.4 \\
\midrule
C & PixArt-$\alpha$ & 13, 50 & 25.7 & 27.1 & 56.9 & 40.8 & 45.3 & 92.4 \\
 & SD3-medium & 12, 50 & 19.0 & 21.7 & 61.3 & 40.8 & 45.3 & 92.7 \\
 & DINOv2-L & 19 & 13.7 & 12.0 & 27.8 & 40.8 & 45.3 & 92.8 \\
\bottomrule

  \end{tabular}
\end{table*}

\paragraph{Scaling the training set.}
\Cref{tab:s-scaling} trains the readouts of Tier B4 on the 33 base training scenes
and on 66 and 132 scenes that add appearance variants, at $t=200$. More scenes improve
accuracy on familiar trajectories but hardly change transfer to ellipses.

\begin{table}[!htbp]
  \centering
  \caption{\textbf{Training-set scaling} (Tier B4, PixArt-$\alpha$, $t=200$). Single-image
  angular error (degrees) on familiar (Fam.) and elliptical (Ell.) trajectories.}
  \label{tab:s-scaling}
  \small
  \setlength{\tabcolsep}{2.5pt}
  \resizebox{\columnwidth}{!}{%
  \begin{tabular}{@{}cc|cc|cc|cc@{}}
    \toprule
    & & \multicolumn{2}{c|}{PixArt-$\alpha$} & \multicolumn{2}{c|}{Geom.\ linear} & \multicolumn{2}{c}{Geom.\ MLP}\\
    Scenes & Block & Fam. & Ell. & Fam. & Ell. & Fam. & Ell.\\
    \midrule
    33 & 14 & 21.3 & 35.9 & 19.8 & 55.3 & 8.3 & 39.9 \\
66 & 14 & 13.8 & 34.0 & 19.1 & 54.7 & 8.0 & 42.5 \\
132 & 15 & 10.6 & 32.0 & 18.0 & 53.4 & 5.1 & 35.5 \\
\bottomrule

  \end{tabular}}
\end{table}

\paragraph{Plane sharing at other blocks.}
\Cref{tab:s-blocks} repeats the comparison of Table 2 at blocks 4 and 14. Lighting
and floor variation leave the sphere plane unchanged at every block, and the frames of
all variants agree with that of spheres to within $11^\circ$, except the colour plane at
block 4 ($20^\circ$).

\begin{table*}[t]
  \centering
  \caption{\textbf{Plane sharing across blocks} (PixArt-$\alpha$, $t=50$). For each block,
  the held-out $R^2$ of the variant's own plane, its subspace overlap with the sphere
  plane, and the rotation of its $(\cos\alpha,\sin\alpha)$ frame relative to spheres
  (degrees).}
  \label{tab:s-blocks}
  \small
  \begin{tabular}{@{}l|ccc|ccc|ccc@{}}
    \toprule
    & \multicolumn{3}{c|}{Block 4} & \multicolumn{3}{c|}{Block 6} & \multicolumn{3}{c}{Block 14}\\
    Variant & $R^2$ & Overlap & Rot. & $R^2$ & Overlap & Rot. & $R^2$ & Overlap & Rot.\\
    \midrule
    A: spheres & 0.38 & 1.00 & 0 & 0.63 & 1.00 & 0 & 0.47 & 1.00 & 0 \\
A: cylinders & 0.28 & 0.72 & -5 & 0.35 & 0.69 & -9 & 0.32 & 0.43 & 1 \\
A: cubes (pair) & 0.25 & 0.63 & -1 & 0.34 & 0.64 & -5 & 0.32 & 0.43 & -1 \\
A: cubes (world) & 0.23 & 0.62 & -3 & 0.32 & 0.64 & -11 & 0.28 & 0.45 & 2 \\
B1: light & 0.37 & 0.90 & -2 & 0.62 & 0.90 & -2 & 0.49 & 0.89 & 0 \\
B2: colour & 0.25 & 0.42 & -20 & 0.52 & 0.48 & 3 & 0.39 & 0.51 & 1 \\
B3: floor & 0.37 & 0.99 & -2 & 0.61 & 0.98 & -2 & 0.45 & 0.97 & 0 \\
\bottomrule

  \end{tabular}
\end{table*}

\paragraph{Shape and appearance results.}
\Cref{tab:s-factors} lists the numbers plotted in Fig.~5 of the main text.

\begin{table*}[t]
  \centering
  \caption{\textbf{Shape and appearance} (Tier A shapes, Tiers B1--B4 and Tier C; 8
  test scenes; plotted in Fig.~5 of the main text). $R^2$: held-out $R^2$ of the
  dataset's own shared plane in PixArt-$\alpha$ (difference features, block 6, $t=50$).
  Single-image angular error ($^\circ$) of the validation-selected readout on training
  paths (Fam.) and on test ellipse views within (In) and outside (Out) the coverage.
  Tier C uses 8-pixel pooling windows. C, per scene: plane and decoders fitted within
  each test scene on its own circles ($R^2$ and Fam.\ on a held-out circle; coverage
  relative to the scene's own cameras).}
  \label{tab:s-factors}
  \small
  \setlength{\tabcolsep}{4pt}
  \begin{tabular}{@{}lc|ccc|ccc|ccc@{}}
    \toprule
    & & \multicolumn{3}{c|}{PixArt-$\alpha$} & \multicolumn{3}{c|}{DINOv2-L} & \multicolumn{3}{c}{Geometry MLP}\\
    Dataset & $R^2$ & Fam. & In & Out & Fam. & In & Out & Fam. & In & Out\\
    \midrule
    A: spheres & 0.63 & 4.6 & 4.5 & 6.5 & 2.6 & 2.6 & 6.8 & 3.6 & 2.4 & 63.4 \\
A: cylinders & 0.35 & 2.6 & 3.0 & 67.8 & 1.5 & 1.2 & 54.6 & 3.0 & 2.6 & 86.5 \\
A: cubes, pair-aligned & 0.34 & 3.3 & 3.5 & 34.3 & 2.0 & 2.3 & 69.5 & 3.7 & 3.6 & 67.6 \\
A: cubes, world-aligned & 0.32 & 3.7 & 4.1 & 9.8 & 2.6 & 2.5 & 31.7 & 4.6 & 4.3 & 79.1 \\
\midrule
B1: light & 0.62 & 5.6 & 6.1 & 8.1 & 2.8 & 2.4 & 6.9 & 4.5 & 3.7 & 67.5 \\
B2: colour & 0.52 & 12.6 & 9.8 & 11.8 & 3.2 & 2.8 & 6.0 & 3.6 & 2.4 & 63.4 \\
B3: floor & 0.61 & 11.0 & 9.7 & 39.8 & 5.2 & 5.6 & 16.7 & 3.6 & 2.4 & 63.4 \\
B4: all varied & 0.31 & 21.2 & 17.1 & 58.5 & 6.3 & 6.5 & 34.3 & 8.3 & 8.2 & 76.5 \\
\midrule
C: all scanned objects & 0.00 & 27.1 & 22.7 & 94.4 & 12.0 & 10.4 & 47.5 & 40.8 & 49.4 & 42.2 \\
C: per scene & 0.24 & 3.5 & 2.9 & 83.5 & 2.4 & 2.2 & 70.3 & 3.3 & 2.6 & 48.9 \\
\bottomrule

  \end{tabular}
\end{table*}

\paragraph{Tier C at matched pooling windows.}
\Cref{tab:s-windows} compares all seven models on Tier C at pooling windows of 8, 16
and 24 pixels (Sec.~4.4 of the main text; procedure in \cref{sec:s-impl}). Fig.~5 of
the main text (\cref{tab:s-factors}) reports the 8-pixel window for PixArt-$\alpha$, DINOv2
and the geometric MLP.

\begin{table}[!htbp]
  \centering
  \caption{\textbf{Tier C: scanned objects at matched pooling windows} (8 test scenes;
  one noise seed, $t=50$ for diffusion models). Plane: best held-out $R^2$ over windows.
  Single-image error ($^\circ$) within the coverage (8- and 24-pixel windows) and outside
  it (8, 16, 24 pixels; 95\% interval at 24).}
  \label{tab:s-windows}
  \small
  \setlength{\tabcolsep}{2.2pt}
  \resizebox{\columnwidth}{!}{%
  \begin{tabular}{@{}lc|cc|ccc@{}}
    \toprule
    & & \multicolumn{2}{c|}{Within} & \multicolumn{3}{c}{Outside}\\
    Model & Plane & 8 & 24 & 8 & 16 & 24\\
    \midrule
    SD 1.5 (UNet) & 0.14 & 6.6 & 6.4 & 14.2 & 14.2 & 14.5 [11, 19] \\
MAE-L & 0.06 & 11.9 & 12.8 & 21.8 & 20.9 & 18.4 [14, 23] \\
DINOv2-L & 0.06 & 10.4 & 11.3 & 47.5 & 40.4 & 28.2 [22, 35] \\
CLIP-L & 0.05 & 24.5 & 20.1 & 56.1 & 48.5 & 35.1 [24, 46] \\
SigLIP & 0.06 & 21.4 & 19.8 & 46.5 & 36.8 & 37.7 [25, 53] \\
PixArt-$\alpha$ & 0.04 & 22.7 & 19.3 & 94.4 & 101.0 & 100.2 [75, 121] \\
SD3-medium & 0.08 & 21.2 & 16.8 & 107.7 & 102.1 & 101.0 [83, 117] \\
\midrule
Geometry MLP & -- & \multicolumn{2}{c}{49.4} & \multicolumn{3}{c}{42.2 [29, 56]} \\
\bottomrule

  \end{tabular}}
\end{table}

\FloatBarrier

\end{document}